\documentclass[sigconf]{acmart}
\renewcommand\footnotetextcopyrightpermission[1]{}
\usepackage{amsmath,amsfonts,bm}

\def\eqref#1{equation~\ref{#1}}
\def\1{\bm{1}}

\DeclareMathAlphabet{\mathsfit}{\encodingdefault}{\sfdefault}{m}{sl}
\SetMathAlphabet{\mathsfit}{bold}{\encodingdefault}{\sfdefault}{bx}{n}
\newcommand{\tens}[1]{\bm{\mathsfit{#1}}}
\def\tA{{\tens{A}}}

\def\sX{{\mathbb{X}}}

\usepackage{amsmath}
\usepackage{amsthm}
\usepackage{wrapfig}
\usepackage{color}
\usepackage{multirow}
\usepackage{subcaption}
\usepackage{cleveref}
\AtBeginDocument{%
  }

\begin{document}

\title{Retrieval-guided Cross-view Image Synthesis}


\author{Hongji Yang }
 \authornote{Both authors contributed equally to this research.}
 \email{2070276033@email.szu.edu.cn} 
 \author{Yiru Li} 
 \authornotemark[1]
 \email{2350271010@email.szu.edu.cn}
 \affiliation{%
  \institution{Shenzhen University}
  \city{Shenzhen}
  \country{China}
}

\author{Yingying Zhu} 
\email{zhuyy@szu.edu.cn}
 \affiliation{%
  \institution{Shenzhen University}
  \city{Shenzhen}
  \country{China}
}

\begin{abstract}
Information retrieval techniques have demonstrated exceptional capabilities in identifying  semantic similarities across diverse domains through robust feature representations. However, their potential in guiding synthesis tasks, particularly cross-view image synthesis, remains underexplored. Cross-view image synthesis presents significant challenges in establishing reliable correspondences between drastically different viewpoints. To address this, we propose a novel retrieval-guided framework that reimagines how retrieval techniques can facilitate effective cross-view image synthesis. Unlike existing methods that rely on auxiliary information, such as semantic segmentation maps or preprocessing modules, our retrieval-guided framework captures semantic similarities across different viewpoints, trained through contrastive learning to create a smooth embedding space.  Furthermore, a novel fusion mechanism leverages these embeddings to guide image synthesis while learning and encoding both view-invariant and view-specific features. To further advance this area, we introduce VIGOR-GEN, a new urban-focused dataset with complex viewpoint variations in real-world scenarios. 
Extensive experiments demonstrate that our retrieval-guided approach significantly outperforms existing methods on the CVUSA, CVACT and VIGOR-GEN datasets, particularly in retrieval accuracy (R@1) and synthesis quality (FID). 
Our work bridges information retrieval and synthesis tasks, offering insights into how retrieval techniques can address complex cross-domain synthesis challenges.
\end{abstract}





\maketitle

\section{Introduction} 

While information retrieval techniques have shown remarkable success in various domains, their potential in guiding image synthesis remains largely unexplored. In this paper, we investigate how retrieval mechanisms can be leveraged to address a fundamental challenge in cross-view image synthesis: bridging the significant domain gap between different viewpoints. 
%We provide a new perspective of retrieval techniques beyond traditional retrieval tasks, suggesting their effectiveness in solving complex visual generation problems.
We demonstrate how advances in retrieval techniques can beyond traditional retrieval scenarios, offering valuable insights for addressing complex cross-domain visual tasks through semantic similarity modeling.

%Although information retrieval techniques have shown remarkable success in various domains, their potential in guiding image synthesis remains largely unexplored. In this paper, we investigate how retrieval mechanisms can be leveraged to address a fundamental challenge in cross-view image synthesis: bridging the significant domain gap between different viewpoints. This work tries to provide a  of retrieval techniques beyond traditional search tasks, demonstrating their effectiveness in solving complex visual generation problems.

Cross-view image synthesis aims to generate an image from a novel viewpoint, given one input image from different viewpoint, such as transforming an aerial (bird's-eye) view into a ground (street) view.  \cite{Pix2Pix, regmi2018cross, tang2019multi, wu2022cross, coming2021,S2SP}. 
% This technique is essential for a wide range of applications, including autonomous driving, robot navigation, 3D reconstruction \cite{mahmud2020boundary}, virtual/augmented reality \cite{bischke2016contextual}, and urban planning \cite{mattyus2017deeproadmapper}.
This technique can benefit a wide range of applications, from autonomous driving and robot navigation to 3D reconstruction \cite{mahmud2020boundary}, virtual/augmented reality \cite{bischke2016contextual}, and urban planning \cite{mattyus2017deeproadmapper}.
\label{sec:intro}
\begin{figure}
    \centering
    \includegraphics[width=0.99\linewidth]{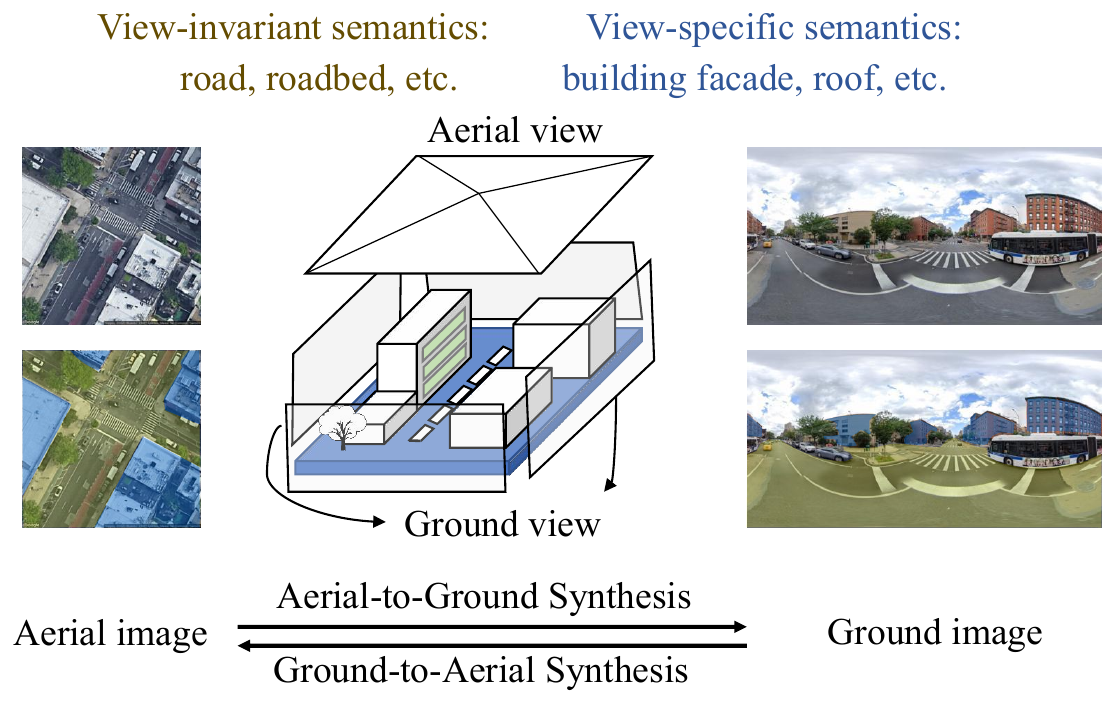}
    \caption{Illustrating \textcolor[RGB]{100,76,0}{view-invariant} and \textcolor[RGB]{47,85,151}{view-specific} semantics in cross-view image synthesis between aerial and ground views.}
   % \caption{Cross-view image synthesis: Illustrating \textcolor[RGB]{100,76,0}{view-invariant semantics} and \textcolor[RGB]{47,85,151}{view-specific semantics} in aerial or ground view .}
    \vspace{-5mm}
\label{fig1}
\end{figure}

While promising, existing cross-view image synthesis methods often rely on auxiliary information, such as semantic segmentation maps \cite{regmi2018cross, tang2019multi, wu2022cross}, or employ preprocessing modules like polar-transformation \cite{Luetal, coming2021, S2SP} to bridge the domain gap between different views. These additional requirements not only introduce significant computational overhead but also complicate the synthesis process, particularly in reverse generation scenarios (e.g., ground-to-aerial synthesis). The reliance on such auxiliary components fails to effectively address a critical challenge in cross-view synthesis - establishing reliable correspondences between drastically different viewpoints, where visual appearances exhibit significant variations. As shown in Figure \ref{fig1}'s lower half, view-specific semantics (highlighted in translucent blue) represent objects with drastically different appearances across viewpoints, exemplified by a building's roof in aerial view versus its facade in ground view. View-specific semantics not only play an important role in establishing reliable correspondences between different viewpoints, but also provide clues to the fidelity and realism of the synthesized images. 
In contrast to view-specific semantics, view-invariant semantics (highlighted in 
translucent yellowish-green) represent elements that preserve their essential 
characteristics across different viewpoints - for example, while roads may appear 
visually different from aerial and ground views, their basic layout remain consistent, enabling reliable cross-view correspondence.
%Correspondingly, view-invariant semantics (highlighted in translucent yellowish green) refer to elements that maintain fundamental similarity across viewpoints despite visual differences, such as roads viewed from aerial and ground views. 
We observe that these challenges parallel key problems in information retrieval, particularly in capturing semantic similarities despite visual differences. This parallel suggests that retrieval techniques could offer valuable insights for addressing cross-view synthesis challenges. 

Moreover, existing datasets for cross-view image synthesis primarily focus on rural and suburban areas, overlooking the complexities of urban environments. This lack of diversity in training data makes it challenging to develop models that can effectively synthesize images in more realistic and challenging scenarios. 

%To tackle these challenges, we propose a new cross-view image synthesis method that does not require semantic segmentation maps or preprocessing modules while generating high-fidelity, realistic target-view images by fully leveraging view-invariant and view-specific semantics. Inspired by the retrieval task's nature of measuring similarity in view-invariant semantics, we introduce a retrieval-guided framework as an embedder to encode these semantics and guide the generation process. This approach obviates the need for preprocessing or segmentation maps for cross-view image pairs.

Inspired by the inherent capability of retrieval tasks to measure semantic similarity, we leverage a retrieval network as an embedder to encode these semantics and guide the generation process, eliminating the need for additional annotations or preprocessing steps. The novel retrieval-guided cross-view image synthesis method that does not require semantic segmentation maps or preprocessing modules while generating high-fidelity, realistic target-view images by fully leveraging view-invariant and view-specific semantics. 
To enhance image quality and realism, our method incorporates view-specific semantics by adopting noise and a modulated style to diversify visual features. We fuse retrieval embedding and style at various layers to improve consistency and image quality. Additionally, to address the scarcity of urban datasets for cross-view image synthesis, we introduce VIGOR-GEN, a derived urban dataset. We validate our proposed method through comprehensive experiments on CVUSA \cite{zhai2017predicting}, CVACT  \cite{liu2019lending}, and the more challenging VIGOR-GEN dataset. Our retrieval-guided model generates more realistic images and significantly outperforms state-of-the-art methods, particularly in terms of SSIM and FID. Extensive ablation studies corroborate the efficacy of each component in our method.

The main contributions of our work are summarized as follows:
\begin{itemize}
    \item \textbf{Introduction of Retrieval Techniques in Cross-view Image Synthesis}. Since retrieval networks have traditionally excelled at measuring semantic similarity, we demonstrate their potential as a powerful guiding mechanism for generative tasks. This dual application of retrieval techniques opens up new possibilities for the retrieval and synthesis research communities.
    \item \textbf{Retrieval-guided Framework for Bridging Domain Gap}. We introduce a retrieval-guided framework that leverages a retrieval network as an embedder. This network is trained to measure the similarity of view-invariant features between different views, effectively bridging the domain gap without needing semantic segmentation maps or preprocessing modules. Our model simplifies the synthesis process and makes reverse generation more straightforward.
    \item \textbf{Novel Generator for Enhanced Semantic Consistency and Diversity}. Our method includes a new generator that incorporates both retrieval embedding and style information at various layers. This approach improves the correspondence between views by leveraging view-invariant semantics captured by the retrieval network while also enhancing the diversity and realism of view-specific semantics using noise and modulated style techniques. This leads to synthesized images with higher fidelity and a more natural appearance.
    \item \textbf{New Dataset for Urban Environments (VIGOR-GEN)}. We build a new derived dataset called VIGOR-GEN, which provides a more challenging and realistic setting for training and evaluating cross-view image synthesis models, pushing the boundaries of the field beyond existing rural and suburban datasets. Our method demonstrates superior performance in synthesizing photo-realistic images from a single input image in another view, as evidenced by its performance on well-known datasets and the new VIGOR-GEN dataset.
\end{itemize}

\section{Related Work}
\label{sec:formatting}

\paragraph{Semantic-guided Cross-view Synthesis}
% X-Fork/X-Seq, SelectionGAN, 
The first pipeline is to apply the semantic segmentation maps of the target-view images to guide the generative model.
% Zhai   propose a linear transformation module to 
Zhai \cite{zhai2017predicting} proposed a linear transformation module to generate a panorama through supervised information from a transformed semantic layout of aerial images.
Regmi and Borji \cite{regmi2018cross} designed two cGAN models, X-fork and X-seq, for simultaneously predicting the target image and the semantic map. 
% In this way, the networks are expected to fully understand the semantic information of the target view and coarsely estimate the semantic correspondence between two views. 
% Tang et. al. propose a two-stage generator for fined-grained image synthesis. 
Tang \cite{tang2019multi} regarded cross-view image synthesis as an image-to-image translation task. This work applied the semantic map of the target view and the source view image as inputs and then obtained the predicted target images. 
% However, this strategy defies the definition of cross-view image synthesis because the semantic maps of target view are often not available in real scenarios. 
% Besides, the information of the target view is leaked to some extent.
To generate 360-degree panorama images, Wu 
 \cite{wu2022cross} proposed PanoGAN as well as a new discrimination mechanism. 
% However, the generated images still lack quality and lean on the semantic maps as supervision.
% However, the quality of such generated images remains unsatisfactory, with reliance on semantic maps for supervision.
Zhu  \cite{ppgan} proposed a Parallel Progressive GAN to stabilize the training of cross-view image synthesis and thus generated rich details.

\paragraph{Preprocessing-guided Cross-view Synthesis}
% Lu, CDE, s2sp
Another pipeline involves a preprocessing module to assimilate the source view image into the target view image.
% Lu  . propose a projection transformation module to estimate the height and semantic information from aerial images. 
% This approach requires the ground-truth height supervision, 
Lu  \cite{Luetal} proposed a projection transformation module that is trained by height and semantic information estimated from aerial images. However, this approach requires ground-truth height supervision for the dataset and carries a complicated pipeline.
Toker  \cite{coming2021} first applied the polar transformation proposed by Shi  \cite{SAFA2019} to cross-view image synthesis, which greatly reduces the domain gap between two views. Besides, Toker  \cite{coming2021} proposed a new multi-tasks framework Coming-Down-to-Earth (CDE) for synthesis, where they postulated that retrieval and synthesis tasks are orthogonal. This approach further improves the correspondence of generation but fails to produce better image detail and quality.
% but fails in image detail and quality.
Shi   \cite{S2SP} proposed an end-to-end network that employs a learnable geographic projection module to learn the projection relationship from the aerial view to the ground view, and then feed the manipulated image into the later generator.
% to obtain further results.
% This explicit or implicit preprocessing of the input image is beneficial for cross-view image synthesis, but it also brings additional computational overhead and the limitation of generating direction. 
% For example, Toker  's method can only generate panoramic images from aerial view images, but not vice versa. 
%For example, CDE applies the algorithm of projecting the aerial view image to the panorama, hence it is difficult to generate the aerial image from the panorama.

% Recently, with the development of the cross-view image geo-localization (essentially retrieval tasks), some approaches \cite{L2LTR2021,zhu2022transgeo,SAIG} have been able to overcome the problem of large domain gap in cross-view image pairs without any auxiliary means.
% This offers the potential to boost the cross-view image synthesis task.

% As a striking difference from existing works, our model is free of semantic maps and preprocessings to realize the mutual generation of ground panorama and aerial image, and retains rich details in the generated target-view images.
%As a striking difference from existing works, in the absence of semantic maps and preprocessing, our model is capable of realizing the mutual generation of ground panorama and aerial image, and retains rich details in the generated target-view images.
As a striking difference from existing works, without the help of semantic maps and preprocessing, our model can synthesize a more realistic target-view image and retain rich details, capable of realizing the mutual generation of ground panorama and aerial image.

\paragraph{Generative Model}
In recent years, diffusion model \cite{rombach2022high, croitoru2023diffusion, ramesh2022hierarchical, saharia2022photorealistic} achieved great success, which produces higher quality images at the cost of a large amount of resources. In addition, there are still neglected problems in cross-view image synthesis, as described in the next section. Moreover, earlier work on cross-view generation does not yield better performance with more artifacts. Therefore, it is essential to study a competitive GAN model before moving fully towards the diffusion model.

\section{Methodology}

% In this section, we first introduce our architecture for cross-view image synthesis.
% % Figure \ref{model_architecture} shows the comparison with other architecture.
% Then, we give an overview of the proposed network in Figure \ref{model_architecture}.
% % Figure \ref{overall_architecture} shows the proposed architecture, which combines several components for specific tasks.

% \subsection{The Mapping Network}

% \subsection{The Retrieval Network}

% \subsection{Generator}

% \subsection{Discriminator}

% \subsection{Loss Function}

\subsection{Overview of Retrieval-Guided Framework}
Inspired by advances in information retrieval, we propose a novel cross-view image synthesis framework that leverages a pre-trained and fixed retrieval model  to identify view-invariant semantics and view-invariant semantics, enabling an end-to-end program without requiring preprocessing or additional input.

% To provide an end-to-end program without the aid of preprocessing or additional input, we propose a new cross-view image synthesis framework that employs a pre-trained and fixed retrieval model as a component to identify the view-invariant semantics in a specific view.
% The embedder trained by contrastive learning can embed the share information into a smooth space and then fuse it in the deep layers. 
% The embedder trained by contrastive learning can embed the view-invariant semantics into a smooth space, and then fuse it in the deep layers. Its insight is to extract an embedding that eliminates visual differences to guarantee the smooth transformation of the view-invariant semantics from source domain to target domain performed by the generator, and thus preserves the structure of the image.

% \begin{wrapfigure}{r}{7cm}
% \vspace{-6mm}
%     \includegraphics[scale=0.4]{arth_new.pdf}
%     \caption{Overview of the proposed framework.} 
%     \vspace{-12mm}
% \label{model_architecture}
% \end{wrapfigure}

% \begin{figure}
%     \centering
%     \includegraphics[width=0.9\linewidth]{arth_new.pdf}
%     \caption{Overview of the proposed framework.}
%     \vspace{-3mm}
%     \label{model_architecture}
% \end{figure}

The embedder, trained through contrastive learning, maps view-invariant semantics into a continuous space, allowing for fusion in the deeper layers. This approach aims to extract embeddings that minimize visual differences, ensuring a smooth transformation of view-invariant semantics from the source domain to the target domain via the generator, thereby preserving the image structure.

% \begin{wrapfigure}{l}
% \includegraphics[scale=0.45]{arth_new.pdf}
% \caption{Overview of the proposed retrieval-guided framework.} 
% \label{model_architecture}
% \end{wrapfigure}
% Its insight is to identity the shared information between two views and make sure that shared representations can be transformed smoothly from source domain to target domain. 
% Its insight is that the shared information between two views can be identified, and generator just needs to focus on how to transform the shared representation from source domain to target domain smoothly.
% Its insight is to ensure the generator focus on how to transform the shared representation from source domain to target domain smoothly. 

% Its insight is to extract an embedding that eliminates visual differences to ensure the generator focuses on how to transform the shared representation from source domain to target domain smoothly. 
Moreover, the embedding can also serve as the condition in the discriminator to guide the generator to improve correspondence.
%which is not feasible in the CDE since it would introduce uncertainty of label.
% without learning to identify the information from scratch.
Meanwhile, we consider the ability of the model to generate view-specific semantics in the target domain by offering modulated style information. 
Although it is difficult to generate identical target-view images, our goal is to ensure that the retrieval-driven view-invariant semantics in the generated images are consistent between the two views while the view-specific semantics remain as visually reasonable as possible.

 \subsection{Network Architecture}\label{sec:network architecture}

\begin{figure*}
\centering
\includegraphics[width = \linewidth]{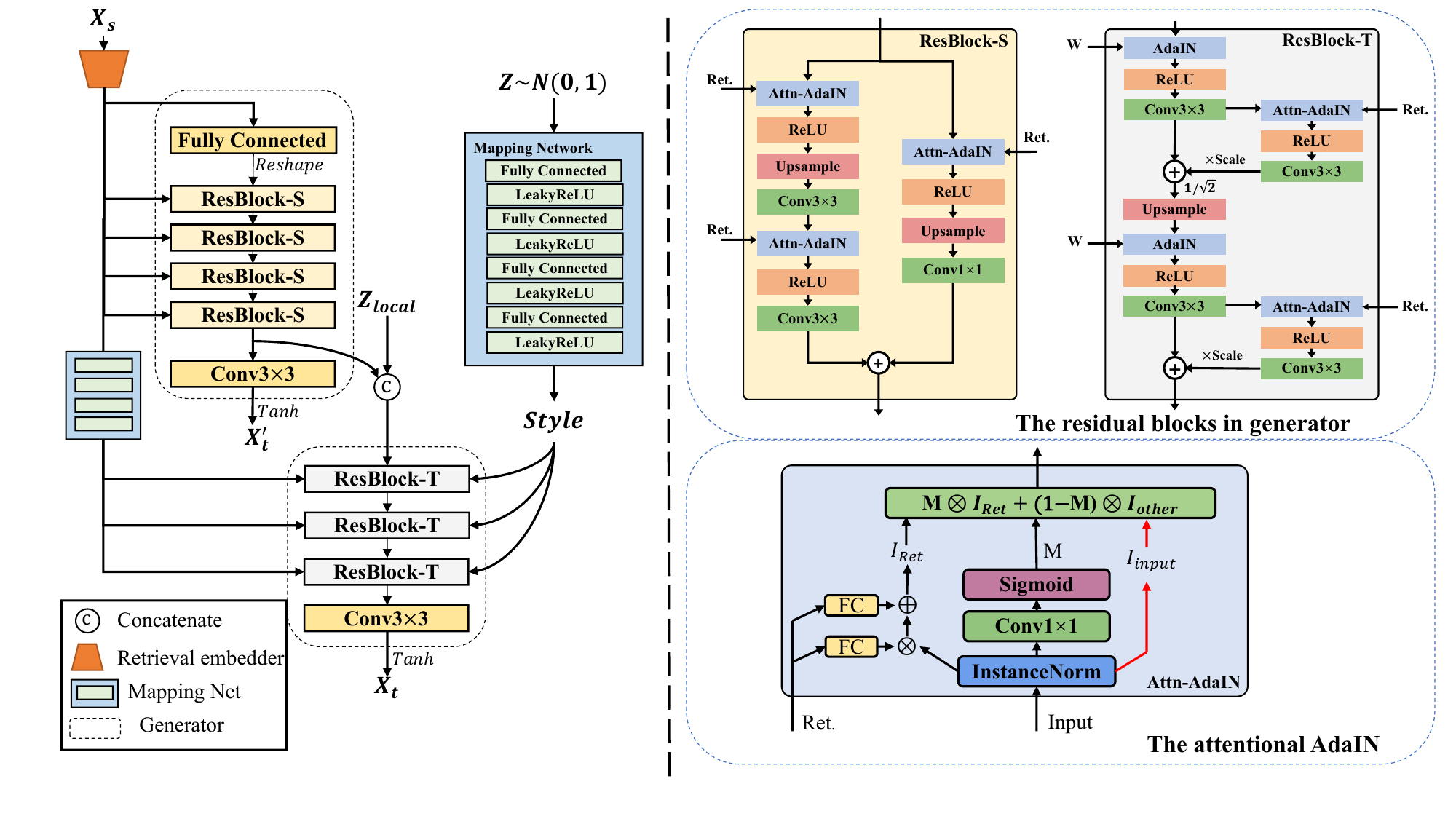}
 \vspace{-10mm}
\caption{\textbf{Illustration of our network architecture.} \textbf{left}: our network consists of a structure generator, a facade generator, a mapping network, and a retrieval embedder. \textbf{right-top}: the residual blocks in our generator. \textbf{right-bottom}: the attentional AdaIN in different residual blocks.
} 
 \vspace{-5mm}
\label{overall_architecture}
\end{figure*}

 The overall architecture of our network is illustrated in Figure \ref{overall_architecture}. It consists of two components: the mapping network and the retrieval network. \textbf{The Mapping Network}: our network has a mapping network which has already been shown in several works 
  \cite{stylegan, stylegan2, karras2020training, choi2020stargan}.
The mapping network learns how to transform the noise sampled from a Gaussian distribution to a new style distribution to better generate view-specific semantics representations, thus yielding detail-enriched images.
This transformation also enhances the quality of image features for retrieval tasks. It consists of four fully connected layers with non-linearity, producing discriminative features useful for retrieval. \textbf{The Retrieval Network}: we adopt the retrieval network proposed in \cite{SAIG} because of its simplicity and effectiveness. It owns stacked attention layers for better feature extraction and encoding for retrieval. The attention mechanism aligns different views, embedding images into a common space for efficient retrieval. We utilize its shallower version SAIG-S \cite{SAIG} here. This retrieval network can settle visual differences and directly embed images from different views into a smooth space.
% Since the architecture of the retrieval network is not the contribution of this work, we will not elaborate on its details here. 
% Please refer to the original paper and the \textcolor[RGB]{0,0,180}{Appendix} of this paper for more details.
% We propose a new framework for cross-view image generation, which uses a pre-trained, fixed retrieval model to extract information that is useful for retrieval between individual views to ensure that shared representations can be mapped smoothly from the source domain to the target domain. At the same time, we take into account the ability of the model to generate unique representations in a particular domain by providing modulated style information to the model.

\subsection{Structure \& Facade Generation}

\paragraph{Two-stage generation}
In general, the generative model controls the generation of structures at low resolutions ($\leq 32\times32$), while features such as facade and color will be affected in higher resolutions ($\geq 32\times32$) \cite{stylegan2, richardson2021encoding, yang2022pastiche}. This hierarchical generation process is highly relevant to information retrieval, as low-resolution structures are first generated and then refined, facilitating a smooth transition in visual features for better matching during retrieval. Therefore, we refine the goals of the generator: at low resolution, the generator focuses on projecting the view-invariant semantics into target-view space. Once the approximate structure of the target view has been generated, the generator then turns its attention to how to generate facades while preserving identity.
% Briefly, we view the task as two subtasks, reconstructing the distribution of shared information in the target view and generating exclusive information.

% We divide the whole generator into two parts: a structure generator for quickly transfer the embedding to the target structure, and a facade generator for generating rich and diverse facades.
\paragraph{Attentional AdaIN}
The embedding extracted by the retrieval model contains the semantic information of the location. Some work \cite{adain, CBN, DF-GAN, park2019semantic, zhu2020sean} has explored how to incorporate the latent code into feature maps to acquire target images. In the context of image retrieval, this latent code helps in transferring the identity and semantic features learned during retrieval directly into the generative process, ensuring that the generated images align with the target view while preserving relevant semantic details. To better inject identity information into the image, we perform some changes to AdaIN \cite{adain} to make feature maps more semantically consistent with the given source image. Given an input $\sX \in\mathbb{R}^{n\times c\times h\times w}$, we first normalize it into zero mean and unit deviation:
% To decide which area should be the shared part in the target view, we utilize an adaptive approach to generate the mask for weighting.

\begin{equation}
\begin{split}
    \hat{\sX}=\frac{\sX-\mu_{nc}}{\sigma_{nc}}, \mu_{nc}=\frac{1}{hw}\sum_{hw}{\sX},
    \sigma_{nc}=\sqrt{\frac{1}{hw}\sum_{hw}({\sX}-\mu_{nc}^2)+\epsilon}
    \end{split}
\end{equation}
where $\epsilon$ is a small constant to prevent the divisor from being zero, $\mu_{nc}$ denotes the mean and $\sigma_{nc}$ denotes the variance. 

Subsequently, the modulation parameters $\gamma$ and $\beta$ are learned by MLP from the retrieval feature $\hat{r}$ :
\begin{equation}
\begin{split}
    \gamma_{r}={MLP}_{\gamma}(\tilde{r}), \beta_{r}={MLP}_{\beta}(\tilde{r}) 
    \end{split}
\end{equation}
Then, the denormalization can be realized as follows:
\begin{equation}
\begin{split}
    \hat{\sX}_r=\gamma_r\hat{\sX}+\beta_{r} 
    \end{split}
\end{equation}

To decide which region and to what extent it can reinforce the retrieval embedding on the image feature, we utilize input $\sX$ to learn to obtain a weight map  $M$. It can be described as:
\begin{equation}
    M=Sigmoid(Conv(\hat{\sX}))
\end{equation}
where $Sigmoid$ denotes the sigmoid activate function. In the ideal case, we expect the modulation of retrieval embeddings to work on the areas where the source view is relevant to the target view. 

Finally, the feature maps are summed by $M$ on the pixel-wise level:
\begin{equation}
    \tilde{\sX} = \hat{\sX}_r \cdot M + \hat{\sX} \cdot (1-M)
\end{equation}

% We argue that the residual structure stabilizes the generation of the target image structure quickly, but introduces more artifacts to the subsequent facade generation. Therefore, the structure generator and the facade generator have different model structures.
% For the generation of the target-view image, we design a structure generator to generate small resolution images (usually $32\times32$), which is used for generating the overall structure. And in the subsequent generation of larger-resolution images, we use a different generator. 

% The structural latent are concatenated with a Gaussian noise for generating exclusive information. The latent feature is then fed into the facade generator. 
% For the structure generator, we use residual structures to make embedding revert to the structure of the target viewpoint image.
\paragraph{Different residual modules}
Residual structures have been widely applied in prior work 
 \cite{regmi2018cross, tang2019multi, wu2022cross, 
 S2SP, ppgan} on cross-view image synthesis to aid in structure generation. However, other work \cite{stylegan, stylegan2} argues that residual structures introduce varying degrees of artifacts and blurring in generation, especially in facade generation. 
Therefore, the modules for generating structures and facades have to be carefully considered, according to different task objectives.

For structure generation, we use a residual structure similar to previous methods, except for the use of an improved AdaIN in the normalization layer.
Both the principal and residual paths are injected with retrieval embedding to facilitate the construction of the structure. 
For facade generation, we follow the network design of previous work \cite{stylegan, stylegan2}, but the residual structure is also used.
The input latent is first fed into AdaIN and the convolution layer to fuse the modulated style. The residual structure is designed to be set after the convolution layer and continue to fuse the embedding through an improved AdaIN. The residual path is then multiplied by a layer scale 
 \cite{touvron2021going, sauer2023stylegan} to perform gradual fading.

\paragraph{Generator}
% Unlike most image-to-image translation or image synthesis tasks, the source and target images in cross-view image synthesis are often of inconsistent resolution, which brings certain constraints to the utilization of the U-net architecture.
% Our generator relies on an encoder-decoder architecture without skip-connection as in U-net. 
% The encoder is the retrieval network with fixed weight and the decoder consists of a set of residual blocks with upsampling. 
% The source image $X_s$ is first fed into the encoder to obtain a retrieval descriptor
% The encoder first gains the retrieval descriptors from the source images $X_s$ and concatenated with a noise vector sampled from Gaussian distribution. 
As shown in Figure \ref{overall_architecture}, our generator first gains the retrieval embedding from the source images $\sX_s$ as the input, which is then integrated into a fully connected layer and is reshaped to be equally proportional to the target image $\sX_t$ in length and width.
The latent feature synthesized by the structure generator is then concatenated with a noise vector sampled from the Gaussian distribution.
The generator gradually increases the scale of the feature map and eventually converts it into an image.
Each residual block in the decoder contains 1) Normalization layers integrating style information or retrieval information; 2) Convolutional layers with spectral normalization \cite{miyato2018spectral} and 3) Activate function.

\paragraph{Discriminator}

To guide our generator to synthesize more realistic and semantically consistent images with the source image, 
we adopt the idea of a one-way discriminator proposed in \cite{DF-GAN}.
% we adopt the one-way discriminator proposed in.
% Please refer to the appendix for specific details about the discriminator.
It first extracts the features of the synthesized image and then concatenates them with the spatially extended embedding vector.
% The discriminator should assign the realistic and matched images at high scores and put the fake or mismatched images at low scores.
The discriminator should assign the realistic and matching images with high scores, and the fake or mismatched images with low scores.
% The details of the discriminator are presented in the \textcolor[RGB]{0,0,180}{Appendix}.

\subsection{Loss Function}

\paragraph{Discriminator Loss}
Since the one-way discriminator is employed, we apply the same adversarial loss \cite{DF-GAN} except for the gradient penalty to train our network.

\begin{equation}
\begin{split}
    \mathcal{L}_{adv}^D=&-\mathbb{E}_{\sX\thicksim\mathbb{P}_r}[\min(0, -1+D(\sX, \tilde{\tA}))] \\
                        &-(1/2)\mathbb{E}_{\hat{\sX}\thicksim\mathbb{P}_g}[\min(0, -1-D(\hat{\sX}, \tilde{\tA}))] \\
                        &-(1/2)\mathbb{E}_{\sX\thicksim\mathbb{P}_{mis}}[\min(0, -1-D(\sX, \tilde{\tA})))]
\end{split}    
\end{equation}
where $\tilde{\tA}$ refers to the retrieval embeddings of the real image $\sX$ and $\hat{\sX}$ denotes the synthesized image.

\paragraph{Generator Loss}
The reconstruction loss is employed to ensure that the target $\sX_t$ is equivalent to the final result $\sX_r$ on a pixel-wise level. 
It can defined as follows:
\begin{equation}
    \mathcal{L}_{rec}=\|\sX_t-\sX_r\|^1
\end{equation}
To further improve the realism, we follow the Learned Perceptual Image Patch Similarity (LPIPS) 
 \cite{LPIPS} loss. Thus, the perceptual loss is defined as:
\begin{equation}
    \mathcal{L}_{perc}=\|\phi(\sX_t)-\phi(\sX_r)\|^1
\end{equation}
where $\phi$ denotes the pre-trained VGG network.

To ensure that the synthesized image has the same view-invariant semantics information as the target image, we use identity loss, which is defined as:
\begin{equation}
\begin{split}
    \mathcal{L}_{id}=& 1-\cos(R(\sX_r),R(\sX_t)) \\
                    &+ 1-\cos(R(\sX^{'}_{r}),R(\sX^{'}_t))
    \end{split}
\end{equation}
where $cos(.,.)$ denotes the cosine similarity between the output embedding vectors and $\sX^{'}_{r}$ means the low-resolution generated image. $R$ denotes the pre-trained retrieval network as in Sec. \ref{sec:network architecture}. 

To prevent the model from generating repetitive content, we apply a diversity loss \cite{mao2019mode, lee2020drit++} between a pair of local code $Z_{local}$. The diversity loss is defined as:
\begin{equation}
    \mathcal{L}_{div}= \frac{d_{z}(z_{{local}_1}, z_{{local}_2}) }{d_{I}(G(w, z_{{local}_1}, \tilde{\tA}), G(w, z_{{local}_2}, \tilde{\tA})) }
\end{equation}
where $d_z(.,.)$ and $d_I(.,.)$ denote the $L1$ distance between the latent codes or images, $G$ is the generator.

\begin{table*}
 \caption{Comparison of VIGOR-GEN with other existing open panorama-aerial cross-view image datasets.}
    \centering
    {\renewcommand\arraystretch{1}
 			\setlength{\tabcolsep}{2.7mm}{
    \begin{tabular}{c|c|c|c|c}
    % \begin{tabular*}{0.95\hsize}{@{}@{\extracolsep{\fill}}cc|c|c|c@{}}
    % {c|c|ccccc|ccccc}
        \bottomrule
       
         Dataset & CVUSA\cite{zhai2017predicting} & CVACT\cite{liu2019lending} & VIGOR\cite{VIGOR2021} & VIGOR-GEN \\ \hline 
         Area & field & suburban & urban & urban \\
         Satellite resolution & $750\times750$ & $1200\times1200$ & $640\times640$ & $640\times640$ \\
         Panorama resolution & $1232\times224$ & $1664\times832$ & $2048\times1024$ & $2048\times1024$ \\
         % Panorama resolution & $1232\times224$ & $1664\times832$ & $2048\times1024$ & $2048\times1024$ \\
         Roughly centered & Yes & Yes & No & Yes \\
         Application & Retrieval, Generation & Retrieval, Generation & Retrieval & Retrieval, Generation \\
         
         % \multirow{2}{*}{Application} & Retrieval, & Retrieval, & \multirow{2}{*}{Retrieval} & \multirow{2}{*}{Generation} \\
         % ~& Generation & Generation & ~ & ~ \\
         \#Satellite Image& 44,416 & 44,416 & 90,618 & 103,516 \\
         \#Panorama Image& 44,416 & 44,416 & 105,214 & 103,516 \\
         \toprule
    \end{tabular}}}
    %\vspace{-9mm}
    \label{tab:datasets}
\end{table*}

The adversarial loss of the generator is as follows:
% \vspace{-0.2mm}
\begin{equation}
    \mathcal{L}_{adv}^G = \mathbb{E}_{\hat{x}\thicksim\mathbb{P}_g}[D(\hat{\sX}, \tilde{\tA})]
\end{equation}

The total loss for the generator is a weighted sum of the above losses, formulated as:
% \vspace{-0.2mm}
\begin{equation}
    \mathcal{L}_{G} = \mathcal{L}^{G}_{adv} + \lambda_{rec}\mathcal{L}_{rec} + \lambda_{perc}\mathcal{L}_{perc} + \lambda_{id}\mathcal{L}_{id} + \lambda_{div}\mathcal{L}_{div}
\end{equation}

\section{VIGOR-GEN Dataset}

For cross-view image synthesis, the commonly used CVUSA \cite{zhai2017predicting} and CVACT 
 \cite{liu2019lending} datasets are primarily field and sub-urban images with an open field of view and less occlusion.
The buildings on both datasets are mostly cottages or bungalows, with simple facade information.
%In contrast to the urban areas with many soaring skyscrapers, this type of dataset is easier in terms of image generation.
% In contrast to the above datasets, the urban areas with soaring skyscrapers contain complicated facades and more occlusions, which is more difficult for image generation.
In contrast to the above datasets, the images with soaring skyscrapers in urban areas often have narrower views and more occlusions, while the complex street surroundings and building facades raise greater challenges to generative networks.
% \paragraph{Urban-Scale Cross-View Image Synthesis dataset}
% Current cross-view image synthesis tasks still suffer from the lack of datasets in urban areas. 
% To this end, we have collected a derived dataset of cross-view urban images, VIGOR-GEN, consisting of 104,585 image pairs.
% However, in real scenarios, cross-view image synthesis is also valuable in urban areas.
% Therefore, to fit real scenarios and to validate the performance of the model in complex scenes, we have collected a derived dataset of cross-view urban images, VIGOR-GEN, consisting of 104,585 image pairs.
To fit realistic scenarios, the cross-view image synthesis generates the need for an urban area dataset.
% Therefore, the cross-view image synthesis raises the need for urban area datasets to fit realistic scenarios.
To this end, we have collected a derived dataset of cross-view urban images, VIGOR-GEN, consisting of 103,516 image pairs.
% The images in urban areas often have narrower views and more occlusions, while the complex street surroundings and building facades raise higher challenges on generative networks.

All images are collected from Google Map API.
The dataset is mainly extended on cross-view image retrieval dataset VIGOR \cite{VIGOR2021}.
To ensure that images synthesized across different views have the same identity as the source image, this task usually requires center-aligned image pairs to avoid ambiguities, so the original VIGOR urban dataset (which is set to be non-centrally aligned) cannot be directly applied to this task.
% Since cross-view image synthesis usually requires center-aligned image pairs, the original VIGOR urban dataset (which is set to be non-centrally aligned) cannot be directly applied to this task.
To extend the application of this dataset, we present a derived dataset in this work so that it can be used for cross-view image synthesis.
Table \ref{tab:datasets} shows the comparison of different datasets.
% \begin{table*}[h]
% \caption{The comparison of VIGOR-GEN and other existing open panorama-aerial cross-view image datasets}
% % \vspace{-2mm}
%     \centering
%     {\renewcommand\arraystretch{1.5}
%  			\setlength{\tabcolsep}{1mm}{
%     \begin{tabular}{c|cccc}
%     % \begin{tabular*}{0.95\hsize}{@{}@{\extracolsep{\fill}}cc|c|c|c@{}}
%     % {c|c|ccccc|ccccc}

%         \bf Dataset & \bf CVUSA & \bf CVACT & \bf VIGOR& \bf VIGOR-GEN 
%          \\ \hline \\
%          Area & field & suburban & urban & urban \\
%          Satellite resolution & $750\times750$ & $1200\times1200$ & $640\times640$ & $640\times640$ \\
%          Panorama resolution & $1232\times224$ & $1664\times832$ & $2048\times1024$ & $2048\times1024$ \\
%          % Panorama resolution & $1232\times224$ & $1664\times832$ & $2048\times1024$ & $2048\times1024$ \\
%          Roughly centered & Yes & Yes & No & Yes \\
%          Application & Retrieval, Generation & Retrieval, Generation & Retrieval & Retrieval, Generation \\
         
%          % \multirow{2}{*}{Application} & Retrieval, & Retrieval, & \multirow{2}{*}{Retrieval} & \multirow{2}{*}{Generation} \\
%          % ~& Generation & Generation & ~ & ~ \\
%          \#Satellite Image& 44,416 & 44,416 & 90,618 & 103,516 \\
%          \#Panorama Image& 44,416 & 44,416 & 105,214 & 103,516 \\
      
%     \end{tabular}}}
%      %\vspace{-6mm}
%     \label{tab:datasets}
% \end{table*}

\section{Experiment}

\subsection{Implementation Details}
\paragraph{Datesets.}
% \paragraph{CVUSA&CVACT}
%We perform our experiments on the panorama-aerial dataset (CVUSA and CVACT), and our newly proposed VIGOR-GEN. 
%Following \cite{coming2021, S2SP}, the CVUSA and CVACT consist of 44,416 image pairs with the train/test split of 35,532/8,884.
%The VIGOR-GEN dataset consists of 51,366 images for training and 51,250 images for testing.
%The resolution of the panorama is set at $128\times512$ in CVUSA and $256\times512$ in both CVACT and VIGOR-GEN. All aerial images are set to a resolution of $256\times256$.
We conducted on three panorama-aerial dataset: CVUSA \cite{CVUSA}, CVACT \cite{liu2019lending} and our proposed VIGOR-GEN datasets. 
Following \cite{coming2021, S2SP}, CVACT \cite{liu2019lending} and CVUSA \cite{CVUSA} contain 35,532 satellite and street-view image pairs for training and 8,884 image pairs for testing.
The VIGOR-GEN dataset comprises 51,366 images for training and 51,250 images for testing.
The resolution of the panorama is set at $128\times512$ in CVUSA and $256\times512$ in both CVACT and VIGOR-GEN. All aerial images are set to a resolution of $256\times256$.

\paragraph{Metrics.}
Following previous work \cite{regmi2018cross, Luetal, coming2021, S2SP}, we adopted the widely used \emph{Structural-Similarity (SSIM)}, \emph{Peak Signal-to-Noise Ratio (PSNR)} and \emph{Learned Perceptual Image Patch Similarity (LPIPS)} \cite{LPIPS} to measure the similarity at the pixel-wise level and feature-wise level, respectively. 
Meanwhile, the realism of the images is measured by \emph{Fr\'echet Inception Distance (FID)} \cite{fid}.
Additionally, we reported the Recall@1 (R@1) in our experiment using another cross-view image retrieval model SAIG-D \cite{SAIG}, which indicates whether the resulting images describe the same location. 
% % We report the Recall@1 (R@1) in our experiment, which indicates whether the first image returned by the retrieval network is correct. 
% % A higher R@1 implies the generated images preserve the identity information better and show a higher correspondence with the target-view image at feature-wise level in terms of retrieval.

\paragraph{Training Details.}
The experiments are implemented using PyTorch. We train our model with 200 epochs using Adam \cite{kingma2014adam} optimizer and $\beta_1=0.5, \beta_2=0.999$. 
The learning rate of the generator and discriminator is set to 0.0001 and 0.0004, respectively.
For each dataset, we use the maximum possible batch size on 4 32GB NVIDIA Tesla V100 GPUs (bs=32 for CVUSA, bs=24 for CVACT, and bs=24 for VIGOR-GEN).
The diversity loss is computed every 4 steps.
We use DiffAug \cite{diffAug} \{Color, Cutout\} as a data augmentation strategy during the training.
The $\lambda_{rec}$, $\lambda_{perc}$ and $\lambda_{id}$ is set to 50, 50 and 10.
The $\lambda_{div}$ is set to 0.1 in CVUSA and CVACT, while is set to 1 in VIGOR-GEN. In previous work, the val set was considered as the test set, note that we only took the final checkpoint for testing and did not select the intermediate checkpoints.

% We train our model with 200 epochs using Adam \cite{kingma2014adam} optimizer and $\beta_1=0.5, \beta_2=0.999$. 
% % The learning rate of the generator is set to 0.0001 and the learning rate of the discriminator is set to 0.0004. 
% The learning rate of the generator and discriminator is set to 0.0001 and 0.0004, respectively.
% % The model is trained with 200 epochs, using 4 V100 GPUs. 
% Please refer to the \textcolor[RGB]{0,0,180}{Appendix} for more details about training.

\subsection{Comparisons with state-of-the-art methods}
We compared our method against several state-of-the-art methods on the CVUSA and CVACT datasets: Pix2Pix \cite{Pix2Pix}, XFork \cite{regmi2018cross}, SelectionGAN \cite{tang2019multi}, PanoGAN \cite{wu2022cross}, CDE \cite{coming2021}, S2SP \cite{S2SP}, PPGAN \cite{ppgan}, Sat2Density \cite{Sat2Density}, ControlNet \cite{ControlNet}, Instruct pix2pix \cite{instructpix2pix},  and CrossViewDiff \cite{croitoru2023diffusion}. 
Results are shown in Table \ref{tab:CVUSAandCVACT} and \ref{tab:VIGOR}. 
As for S2SP \cite{S2SP}, it applies the geometry project equation to calculate the projection from satellite image to street-view panorama, whose inverse process is not given in the original paper, so this method will not be compared at ground-to-aerial (g2a) generation.

\begin{table*}
 % \vspace{-4mm}
\caption{Comparison of competitive methods on CVUSA and CVACT. Note that for the FoV-only model, we follow \cite{tang2019multi} and obtain the final panorama, which consists of four street images with a FoV of 90 degrees. For fair comparison, the semantic map are discarded as an input to SelectionGAN.}
%\vspace{-1mm}
% \scriptsize
% \resizebox{\linewidth}{22mm}{
    \centering
    % \begin{tabular*}{0.9\hsize}{@{}@{\extracolsep{\fill}}c|c|ccccc|ccccc@{}}
     {\renewcommand\arraystretch{1.0}
 			\setlength{\tabcolsep}{2mm}{
    \begin{tabular}{c|l|ccccc|ccccc}
        \bottomrule
         \multirow{2}{*}{\bf Direction} & \multirow{2}{*}{\bf Method} & \multicolumn{5}{c|}{\bf CVUSA} & \multicolumn{5}{c}{\bf CVACT}  \\\cline{3-12}
         ~&~& \bf SSIM$\uparrow$& \bf PSNR$\uparrow$&\bf LPIPS$\downarrow$&\bf FID$\downarrow$&\bf R@1$\uparrow$&\bf SSIM$\uparrow$&\bf PSNR$\uparrow$& \bf LPIPS$\downarrow$ & \bf FID$\downarrow$ & \bf R@1$\uparrow$ \\ \hline
         \multirow{6}{*}{a2g} & Pix2Pix & 0.2849 & 12.14 & 0.5712 & 82.84 & 0.01 & 0.3634 & 13.37 & 0.4943 & 86.21 & 0.00 \\
         ~ & XFork & 0.3408 & 13.25 & 0.5611 & 79.75 & 6.41 & 0.3701 & 14.17 & 0.4919 & 47.98 & 8.72 \\
         ~ & SelectionGAN & 0.3278 & 13.37 & 0.5331 & 90.72 & 4.58 & 0.4705 & 14.31 & 0.5141 & 95.67 & 6.67 \\
         ~ & PanoGAN & 0.3024 & 13.67 & 0.4684 & 75.24 & 33.11 & 0.4631 &  14.18& 0.4762 & 82.65 & 28.71 \\
         ~ & CDE & 0.2980 & 13.87 & 0.4752 & 20.63 & 85.04 & 0.4506 & 13.98 & 0.4927 & 43.96 & 65.04 \\
         ~ & S2SP & 0.3437 & 13.32 & 0.4688 & 44.15 &  10.09 & 0.4521 & 14.14 & 0.4718 & 39.64 & 29.39 \\
         ~ & PPGAN & 0.3516 & 13.91 & - & - & - & - & -& - & - & - \\
         ~ & Sat2Density & 0.3390 & 14.23 & - & 41.43 &- & 0.3870 &14.27 &-&47.09&-\\
        ~ & ControlNet & 0.2770 & 11.18 & - & 44.63 &- &0.3400 & 12.15 &-&47.15 &-\\
         ~ & Instruct pix2pix &0.2550 & 10.66 &-& 68.75 &- &0.3920 & 13.12&- &57.74&-\\
         ~ & CrossViewDiff &\textbf{0.3710} & 12.00 &-&23.67&-&0.4120 &12.41 &-&41.94&-\\
         ~ & Ours & 0.3706 & \textbf{14.33} & \textbf{0.4302} & \textbf{13.57} & \textbf{96.25} & \textbf{0.4945} & \textbf{14.55} & \textbf{0.4540} & \textbf{21.83} & \textbf{87.90} \\\hline
         \multirow{3}{*}{g2a} & Pix2Pix & 0.1956  & 15.07 & 0.6220 & 121.95 & 7.85 & 0.0870 & 14.24 & 0.6612 & 133.39 & 13.06 \\
         ~ & CDE & 0.2167 & 15.19 & 0.5706 & 121.98 & 14.73 & 0.0906 & 14.59 & 0.6689 & 160.81 & 14.99 \\
         ~ & Ours & \textbf{0.2461} & \textbf{15.77} & \textbf{0.5181} & \textbf{41.65} & \textbf{95.14} & \textbf{0.1966} & \textbf{16.29} & \textbf{0.5551} & \textbf{36.54} & \textbf{87.81} \\ 
          \toprule
         
    \end{tabular}}}
    %\vspace{-4mm}
    \label{tab:CVUSAandCVACT}
\end{table*}

\begin{table}
     \caption{Comparison of existing competitive methods on our newly proposed VIGOR-GEN.}
    \centering
    {\renewcommand\arraystretch{1}
 			\setlength{\tabcolsep}{0.5mm}{
    \begin{tabular}{c|l|ccccc}
    % {c|c|ccccc|ccccc}
        \bottomrule
         \multirow{2}{*}{Direction} & \multirow{2}{*}{Method} & \multicolumn{5}{c}{VIGOR-GEN} \\ \cline{3-7}
         ~&~&{SSIM$\uparrow$}&{PSNR$\uparrow$}&{LPIPS$\downarrow$}&{FID$\downarrow$}&{R@1$\uparrow$} \\\hline
            %~&~&SSIM$\uparrow$&PSNR$\uparrow$&LPIPS$\downarrow$&FID$\downarrow$&R@1$\uparrow$ \\\midrule
         \multirow{4}{*}{a2g} & Pix2Pix\cite{Pix2Pix} & 0.3566 & 12.18 & 0.6114 & 100.25 & 0.01  \\
         ~ & SelectionGAN\cite{tang2019multi} & 0.3986 & 13.16 & 0.5234 & 104.22 & 7.41 \\
         ~ & PanoGAN\cite{wu2022cross} & 0.4031 & 13.83 & 0.5467 & 75.76 & 8.49 \\
         ~ & CDE\cite{coming2021} & 0.3672 & 12.72 & 0.6108 & 78.26 & 0.22 \\
         ~ & S2SP\cite{S2SP} & 0.4041 & 13.73 & 0.5422 & 69.28 & 4.54  \\
         ~ & Ours& \textbf{0.4243} & \textbf{13.91} & \textbf{0.4548} & \textbf{13.64} & \textbf{37.94} \\\hline
         \multirow{3}{*}{g2a} & Pix2Pix\cite{Pix2Pix} & 0.1885 & 13.31 & 0.5876 & 96.26 & 2.41  \\
         ~ & CDE\cite{coming2021} & 0.1830 & 12.89 & 0.5734 & 95.13 & 3.25 \\
         ~ & Ours & \textbf{0.1901} & \textbf{13.99} &\textbf{0.5278} & \textbf{30.93} & \textbf{34.58} \\
         % ~ & Ours & 0.1796 & 13.32 & 0.5270 & 21.61 &  \\
         \toprule
    \end{tabular}}}
    %\vspace{-6mm}
    \label{tab:VIGOR}
\end{table}

\paragraph{Quantitative Results}

In aerial-to-ground synthesis, our method demonstrates superior performance across multiple metrics on the CVUSA dataset: surpassing S2SP \cite{S2SP} by 6 points in SSIM and achieving 
improvements of 1.01 and 0.0386 in PSNR and LPIPS, respectively. 
%And it outperforms the state-of-the-art methods CrossViewDiff \cite{croitoru2023diffusion} by 2.33 points in terms of PSNR on the CVUSA dataset and 2.14 points in terms of PSNR on the CVACT dataset and gains an important improvement by 0.0825 points in terms of SSIM on the CVACT dataset.
Compared to the current state-of-the-art CrossViewDiff \cite{croitoru2023diffusion}, our method achieves higher PSNR 
scores with improvements of 2.33 points on CVUSA and 2.14 points on CVACT, while also gaining a 0.0825 point improvement in SSIM on CVACT.
%In ground-to-aerial image synthesis, compared to the most competitive method CDE \cite{coming2021}, our model gains a significant improvement in LPIPS (0.5181 versus 0.5706 on CVUSA), which proves that generated images are more consistent with human visual perception.
%This is attributed to the embedding can be used as the condition on the discriminator to guide the generator to improve the correspondence, which does not apply to CDE \cite{coming2021} as it introduces labeling uncertainty.
For ground-to-aerial synthesis, our model significantly outperforms the previous best method 
CDE \cite{coming2021} in LPIPS (0.5181 vs 0.5706 on CVUSA), indicating better alignment with human visual perception. This improvement stems from our discriminator's ability to leverage embeddings as conditions to guide generation - an advantage over CDE \cite{coming2021}, which suffers from labeling uncertainty.

% realism
%It is worth noting that our method has a larger improvement in FID compared to other models. 
%For example, our method gains a $7.06$ point improvement compared to CDE \cite{coming2021}.
%This is because we consider not only the view-invariant semantics across views but also the view-specific semantics of the target view, which makes the synthesized images more realistic.
%Especially, in ground-to-aerial image synthesis, it is challenging for other models to generate the obscured parts, resulting in a decrease in realism. A lower FID can be observed on CVUSA (41.65 versus 121.95).
Notably, our method achieves substantial improvements in FID scores, with a 7.06-point improvement over CDE \cite{coming2021}. This superior performance stems from our comprehensive approach 
that considers both view-invariant and view-specific semantics, enhancing image realism.
This advantage is particularly evident in ground-to-aerial synthesis, where other models struggle with generating obscured regions. Our method demonstrates significantly better realism on CVUSA, achieving an FID of 41.65 compared to 121.95 for baseline methods.

%Experiments are also conducted on our newly proposed urban dataset VIGOR-GEN which is more challenging due to its complex facades and inevitable occlusions. As a result, the exclusive information in one view is more complicated in an urban setting.
%As shown in Table \ref{tab:VIGOR}, our method outperforms other methods in all metrics. 
%For example, our proposed method sets the new state-of-the-art FID of 13.64 at Aerial-to-Ground (a2g) and 30.93 at g2a on VIGOR-GEN while other methods have a higher FID.
% Our method alleviates the domain gap due to the different views and the limitations on model architecture due to the different resolutions.
%For the R@1 metric, the multi-task framework CDE, which performs well on CVUSA and CVACT, almost fails in VIGOR-GEN. 
%In other words, CDE does not fit well in urban areas while our method still produces images with higher quality. 

\begin{figure*}
%\vspace{-3mm}
        \centering
		\begin{minipage}[t]{0.072\linewidth}
			\centering
			\raisebox{-0.15cm}{\includegraphics[width=1\linewidth]{ 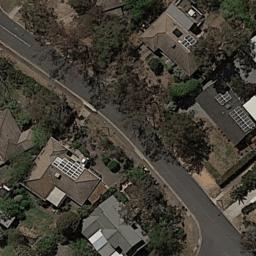}}
		\end{minipage}
		\begin{minipage}[t]{0.144\linewidth}
			\centering
			\raisebox{-0.15cm}{\includegraphics[height=0.5\linewidth,width=1\linewidth]{ 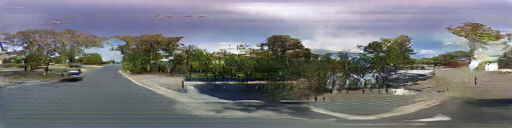}}
		\end{minipage}
        \begin{minipage}[t]{0.144\linewidth}
			\centering
			\raisebox{-0.15cm}{\includegraphics[height=0.5\linewidth,width=1\linewidth]{ 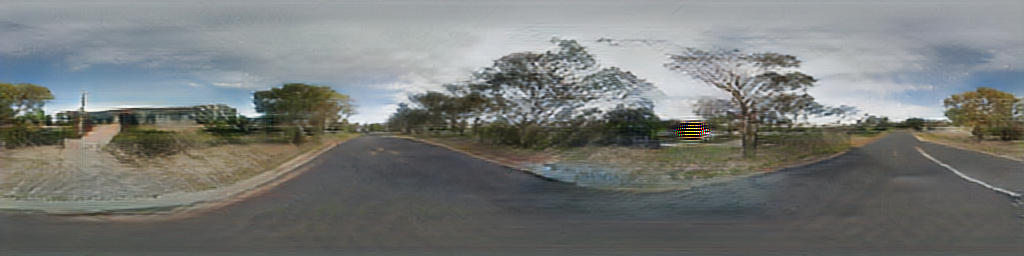}}
		\end{minipage}
		\begin{minipage}[t]{0.144\linewidth}
			\centering
			\raisebox{-0.15cm}{\includegraphics[height=0.5\linewidth,width=1\linewidth]{ 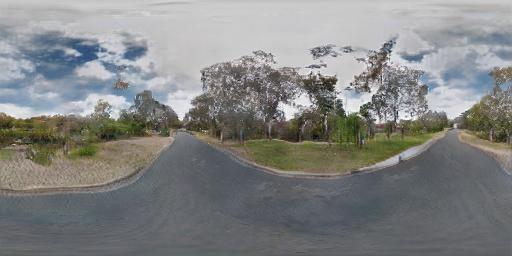}}
		\end{minipage}
		\begin{minipage}[t]{0.144\linewidth}
			\centering
			\raisebox{-0.15cm}{\includegraphics[width=1\linewidth]{ CVACT-1-fake-s2sp.jpg}}
		\end{minipage}
		\begin{minipage}[t]{0.144\linewidth}
			\centering
			\raisebox{-0.15cm}{\includegraphics[width=1\linewidth]{ 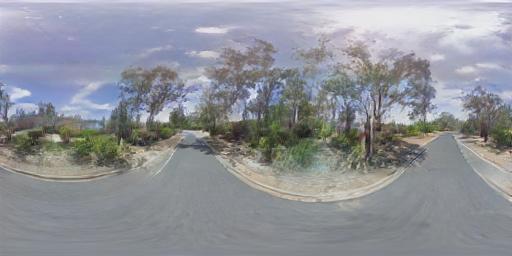}}
		\end{minipage}
        \begin{minipage}[t]{0.144\linewidth}
			\centering
			\raisebox{-0.15cm}{\includegraphics[width=1\linewidth]{ 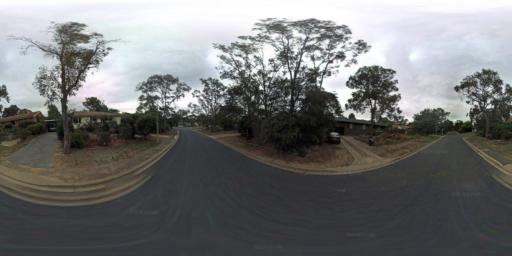}}
		\end{minipage}

 \centering
		\begin{minipage}[t]{0.072\linewidth}
			\centering
			\raisebox{-0.15cm}{\includegraphics[width=1\linewidth]{ 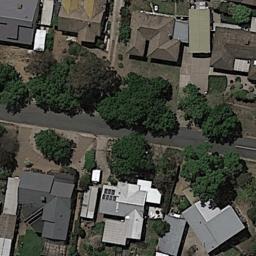}}
		\end{minipage}
		\begin{minipage}[t]{0.144\linewidth}
			\centering
			\raisebox{-0.15cm}{\includegraphics[height=0.5\linewidth,width=1\linewidth]{ 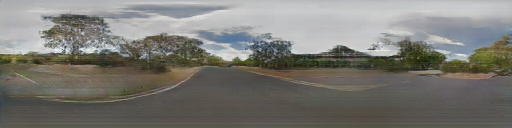}}
		\end{minipage}
            \begin{minipage}[t]{0.144\linewidth}
			\centering
			\raisebox{-0.15cm}{\includegraphics[height=0.5\linewidth,width=1\linewidth]{ 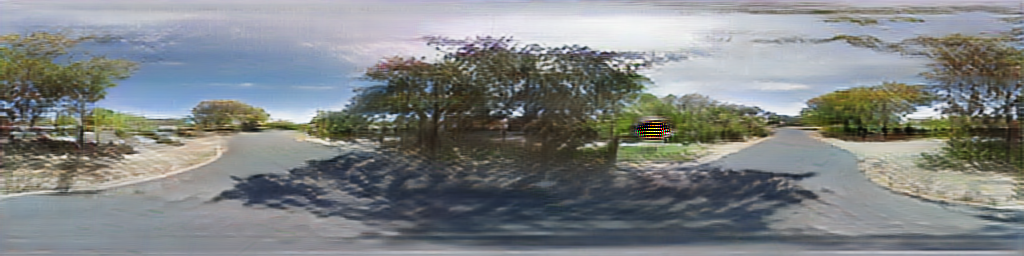}}
		\end{minipage}
		\begin{minipage}[t]{0.144\linewidth}
			\centering
			\raisebox{-0.15cm}{\includegraphics[height=0.5\linewidth,width=1\linewidth]{ 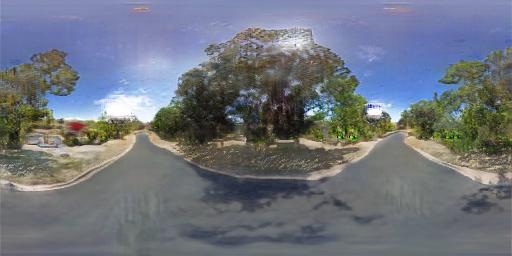}}
		\end{minipage}
		\begin{minipage}[t]{0.144\linewidth}
			\centering
			\raisebox{-0.15cm}{\includegraphics[width=1\linewidth]{ 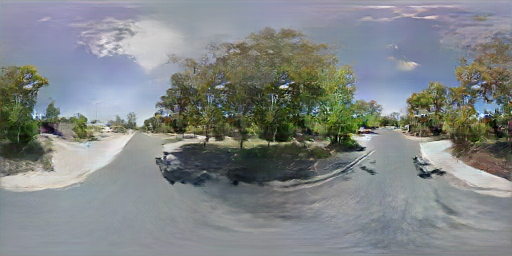}}
		\end{minipage}
		\begin{minipage}[t]{0.144\linewidth}
			\centering
			\raisebox{-0.15cm}{\includegraphics[width=1\linewidth]{ 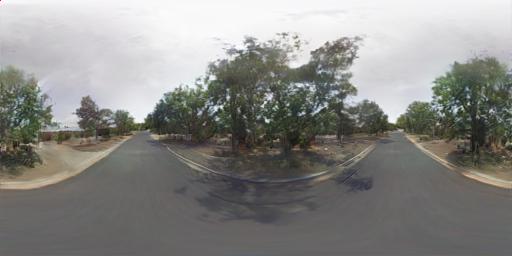}}
		\end{minipage}
        \begin{minipage}[t]{0.144\linewidth}
			\centering
			\raisebox{-0.15cm}{\includegraphics[width=1\linewidth]{ 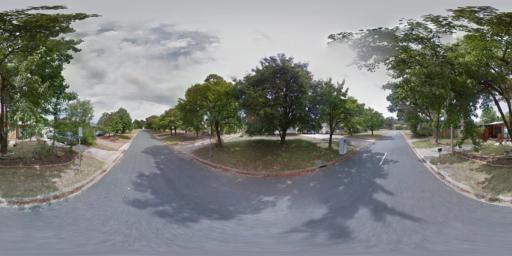}}
		\end{minipage}

    \centering
		\begin{minipage}[t]{0.072\linewidth}
			\centering
			\raisebox{-0.15cm}{\includegraphics[width=1\linewidth]{ 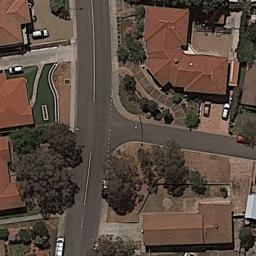}}
		\end{minipage}
		\begin{minipage}[t]{0.144\linewidth}
			\centering
			\raisebox{-0.15cm}{\includegraphics[height=0.5\linewidth,width=1\linewidth]{ 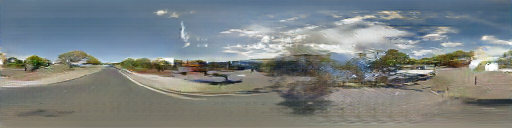}}
		\end{minipage}
            \begin{minipage}[t]{0.144\linewidth}
			\centering
			\raisebox{-0.15cm}{\includegraphics[height=0.5\linewidth,width=1\linewidth]{ 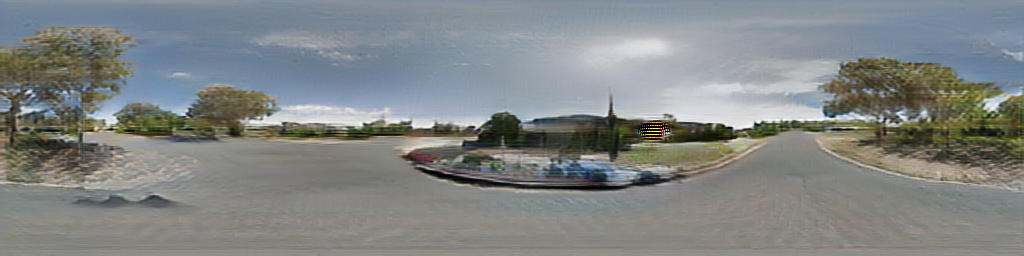}}
		\end{minipage}
		\begin{minipage}[t]{0.144\linewidth}
			\centering
			\raisebox{-0.15cm}{\includegraphics[height=0.5\linewidth,width=1\linewidth]{ 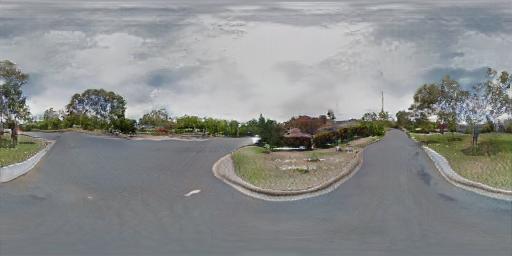}}
		\end{minipage}
		\begin{minipage}[t]{0.144\linewidth}
			\centering
			\raisebox{-0.15cm}{\includegraphics[width=1\linewidth]{ 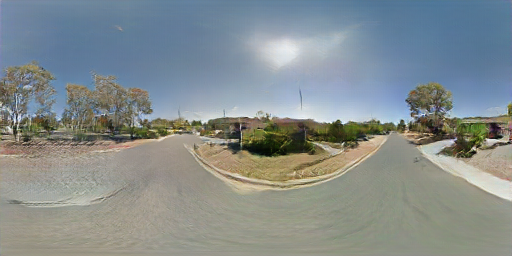}}
		\end{minipage}
		\begin{minipage}[t]{0.144\linewidth}
			\centering
			\raisebox{-0.15cm}{\includegraphics[width=1\linewidth]{ 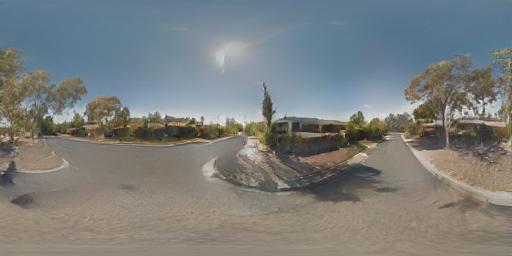}}
		\end{minipage}
        \begin{minipage}[t]{0.144\linewidth}
			\centering
			\raisebox{-0.15cm}{\includegraphics[width=1\linewidth]{ 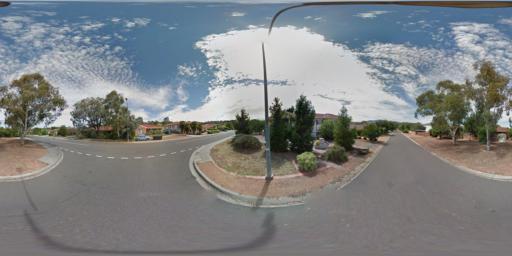}}
		\end{minipage}

    \centering
		\begin{minipage}[t]{0.072\linewidth}
			\centering
			\raisebox{-0.15cm}{\includegraphics[width=1\linewidth]{ 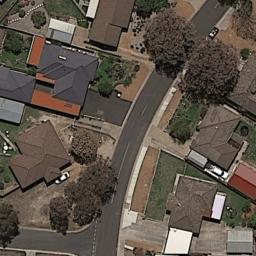}}
		\end{minipage}
		\begin{minipage}[t]{0.144\linewidth}
			\centering
			\raisebox{-0.15cm}{\includegraphics[height=0.5\linewidth,width=1\linewidth]{ 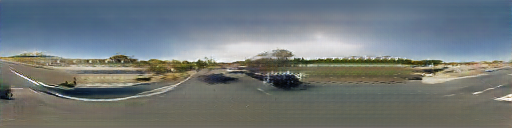}}
		\end{minipage}
            \begin{minipage}[t]{0.144\linewidth}
			\centering
			\raisebox{-0.15cm}{\includegraphics[height=0.5\linewidth,width=1\linewidth]{ 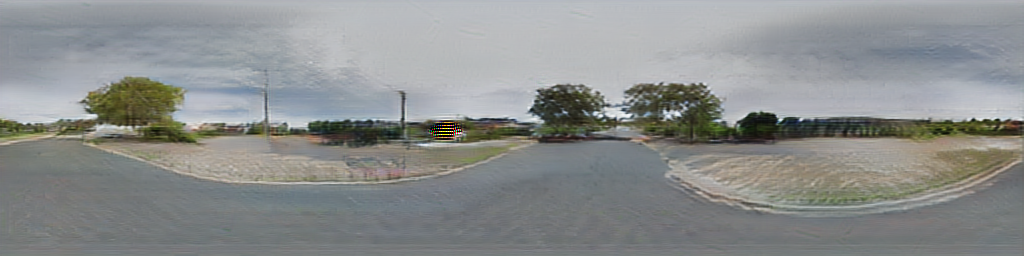}}
		\end{minipage}
		\begin{minipage}[t]{0.144\linewidth}
			\centering
			\raisebox{-0.15cm}{\includegraphics[height=0.5\linewidth,width=1\linewidth]{ 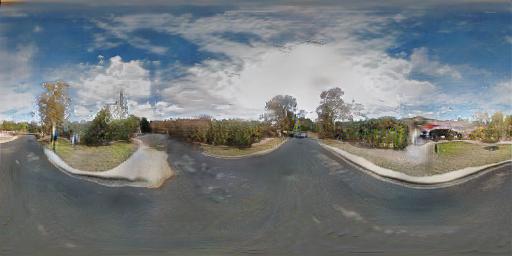}}
		\end{minipage}
		\begin{minipage}[t]{0.144\linewidth}
			\centering
			\raisebox{-0.15cm}{\includegraphics[width=1\linewidth]{ 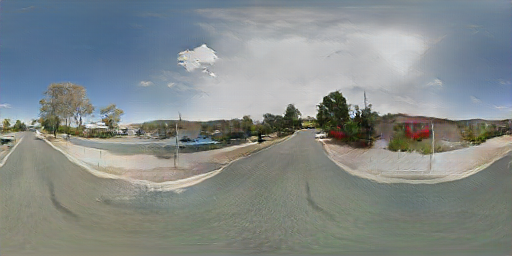}}
		\end{minipage}
		\begin{minipage}[t]{0.144\linewidth}
			\centering
			\raisebox{-0.15cm}{\includegraphics[width=1\linewidth]{ 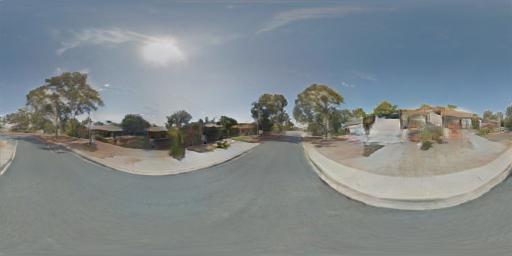}}
		\end{minipage}
        \begin{minipage}[t]{0.144\linewidth}
			\centering
			\raisebox{-0.15cm}{\includegraphics[width=1\linewidth]{ 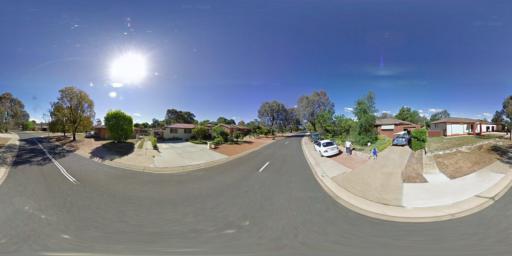}}
		\end{minipage}

    \setcounter{subfigure}{0}
    \centering
	\begin{subfigure}[t]{0.072\linewidth}
		\begin{minipage}[t]{\linewidth}
			\centering
			\raisebox{-0.15cm}{\includegraphics[width=1\linewidth]{ 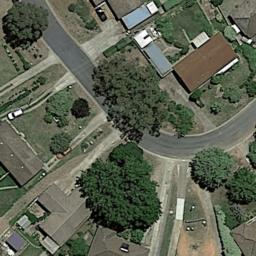}}
		\end{minipage}
        \caption{Input}
        \end{subfigure}
	\begin{subfigure}[t]{0.144\linewidth}
		\begin{minipage}[t]{\linewidth}
			\centering
			\raisebox{-0.15cm}{\includegraphics[height=0.5\linewidth,width=1\linewidth]{ 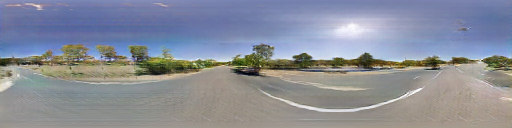}}
		\end{minipage}
        \caption{Pix2Pix}
        \end{subfigure}
        \begin{subfigure}[t]{0.144\linewidth}	
		\begin{minipage}[t]{\linewidth}
			\centering
			\raisebox{-0.15cm}{\includegraphics[height=0.5\linewidth,width=1\linewidth]{ 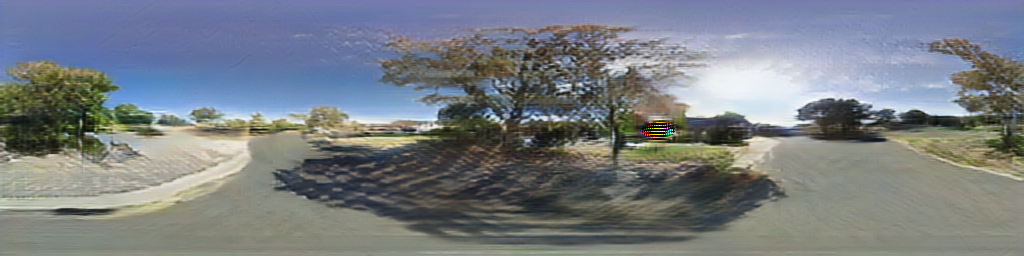}}
		\end{minipage}
        \caption{PanoGAN}
        \end{subfigure}
	\begin{subfigure}[t]{0.144\linewidth}	
		\begin{minipage}[t]{\linewidth}
			\centering
			\raisebox{-0.15cm}{\includegraphics[height=0.5\linewidth,width=1\linewidth]{ 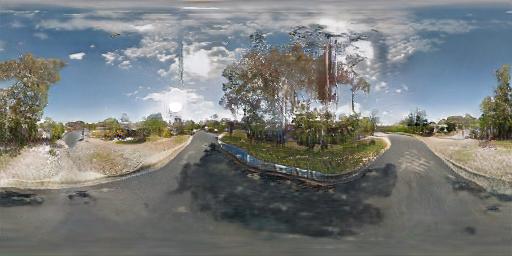}}
		\end{minipage}
        \caption{CDE}
        \end{subfigure}
	\begin{subfigure}[t]{0.144\linewidth}
		\begin{minipage}[t]{\linewidth}
			\centering
			\raisebox{-0.15cm}{\includegraphics[width=1\linewidth]{ 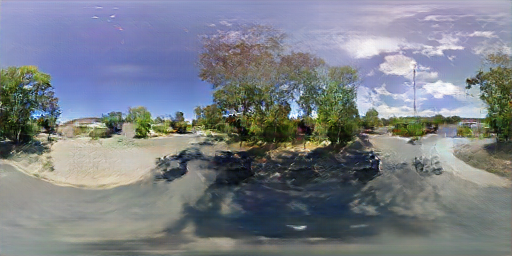}}
		\end{minipage}
        \caption{S2SP}
        \end{subfigure}
	\begin{subfigure}[t]{0.144\linewidth}
		\begin{minipage}[t]{\linewidth}
			\centering
			\raisebox{-0.15cm}{\includegraphics[width=1\linewidth]{ 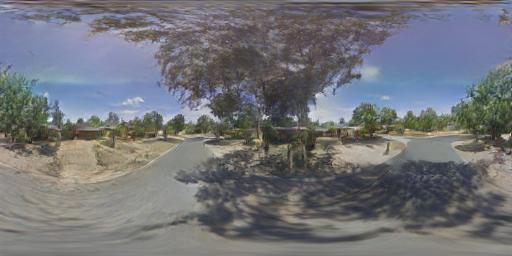}}
		\end{minipage}
        \caption{Ours}
        \end{subfigure}
	\begin{subfigure}[t]{0.144\linewidth}
        \begin{minipage}[t]{\linewidth}
			\centering
			\raisebox{-0.15cm}{\includegraphics[width=1\linewidth]{ 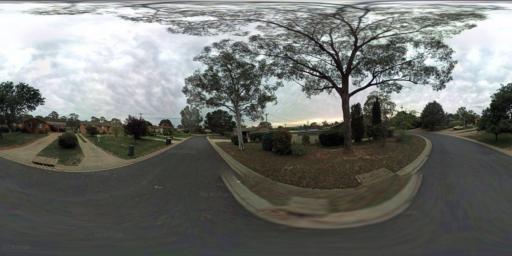}}
		\end{minipage}
        \caption{GT}
        \end{subfigure} 
        %\vspace{-3mm}
	\caption{Comparison of current methods at a2g direction on CVACT.} 
 %\vspace{-7mm}
	\label{CVACT-GAN}
\end{figure*}

We further evaluate our method on the VIGOR-GEN dataset, which presents additional challenges through complex urban facades and occlusions, resulting in more complicated view-specific information.
Our method achieves superior performance across all metrics on VIGOR-GEN (Table \ref{tab:VIGOR}), setting new state-of-the-art FID scores of 13.64 for aerial-to-ground and 30.93 for ground-to-aerial synthesis. 
Notably, while CDE performs well on CVUSA and CVACT, it struggles significantly on VIGOR-GEN's urban scenes, particularly in R@1 metrics. CDE does not fit well in urban areas while our method still generates images with higher quality, suggesting our method's robustness in handling complex urban environments.
\paragraph{Qualitative Results}
We provide the qualitative results of our method on different datasets to demonstrate its effectiveness.
% a2g
As shown in Figure \ref{CVACT-GAN}, our method generates more realistic and detailed images with fewer artifacts compared to existing methods. 

In the first group of Figure \ref{CVACT-GAN}, our approach generates consistent and clear roads with fewer artifacts on CVACT, showcasing its ability to overcome significant visual differences. In addition, our method exhibits exceptional performance in complex scenes. For example,in the first row of the second group of Figure \ref{CVACT-GAN}, our method synthesizes more realistic building facades, including intricate details such as windows and doors. In contrast, other methods fail to reproduce these distinctive features in panoramic views. This superior performance is attributed to our model’s consideration of both the correspondence between the source and target views and the content discrepancies between them. Unlike other models that struggle to handle view-specific semantics information differences, our method performs equally well in urban environments. 

\begin{figure}
%\vspace{-5mm}
    \centering
	\begin{subfigure}[t]{0.32\linewidth}
		\begin{minipage}[t]{\linewidth}
			\centering
			\raisebox{-0.15cm}{\includegraphics[width=1\linewidth]{ 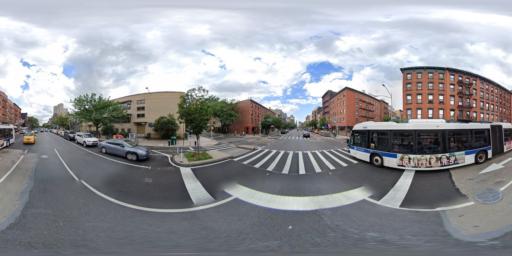}}
		\end{minipage}\hspace{-1mm}
	\end{subfigure}
        \begin{subfigure}[t]{0.16\linewidth}
		\begin{minipage}[t]{\linewidth}
			\centering
			\raisebox{-0.15cm}{\includegraphics[width=1\linewidth]{ 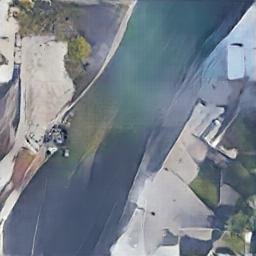}}
		\end{minipage}\hspace{-1mm}
	\end{subfigure}
        \begin{subfigure}[t]{0.16\linewidth}
		\begin{minipage}[t]{\linewidth}
			\centering
			\raisebox{-0.15cm}{\includegraphics[width=1\linewidth]{ 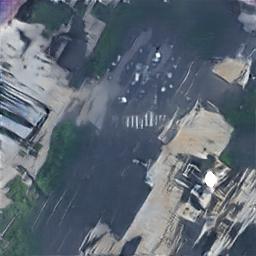}}
		\end{minipage}\hspace{-1mm}
	\end{subfigure}
        \begin{subfigure}[t]{0.16\linewidth}
		\begin{minipage}[t]{\linewidth}
			\centering
			\raisebox{-0.15cm}{\includegraphics[width=1\linewidth]{ 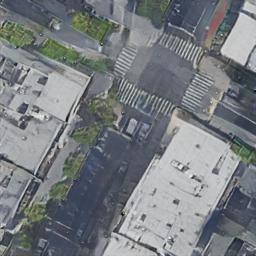}}
		\end{minipage}\hspace{-1mm}
        \end{subfigure}
        \begin{subfigure}[t]{0.16\linewidth}
        \begin{minipage}[t]{\linewidth}
			\centering
			\raisebox{-0.15cm}{\includegraphics[width=1\linewidth]{ 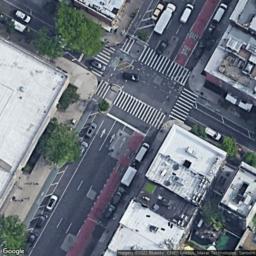}}
		\end{minipage}\vspace{-3mm}\hspace{-1mm}
        \end{subfigure}

    \centering
	\begin{subfigure}[t]{0.32\linewidth}
		\begin{minipage}[t]{\linewidth}
			\centering
			\raisebox{-0.15cm}{\includegraphics[width=1\linewidth]{ 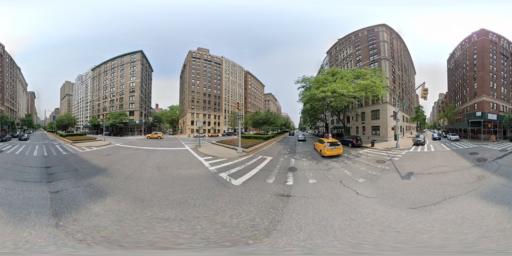}}
		\end{minipage}\hspace{-1mm}
	\end{subfigure}
        \begin{subfigure}[t]{0.16\linewidth}
		\begin{minipage}[t]{\linewidth}
			\centering
			\raisebox{-0.15cm}{\includegraphics[width=1\linewidth]{ 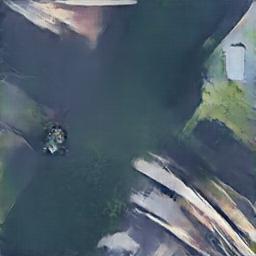}}
		\end{minipage}\hspace{-1mm}
	\end{subfigure}
        \begin{subfigure}[t]{0.16\linewidth}
		\begin{minipage}[t]{\linewidth}
			\centering
			\raisebox{-0.15cm}{\includegraphics[width=1\linewidth]{ 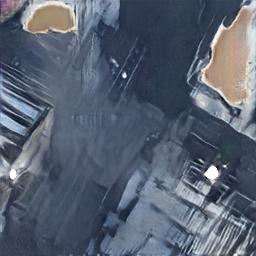}}
		\end{minipage}\hspace{-1mm}
	\end{subfigure}
        \begin{subfigure}[t]{0.16\linewidth}
		\begin{minipage}[t]{\linewidth}
			\centering
			\raisebox{-0.15cm}{\includegraphics[width=1\linewidth]{ 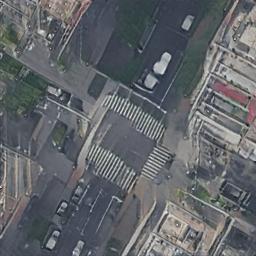}}
		\end{minipage}\hspace{-1mm}
        \end{subfigure}
        \begin{subfigure}[t]{0.16\linewidth}
        \begin{minipage}[t]{\linewidth}
			\centering
			\raisebox{-0.15cm}{\includegraphics[width=1\linewidth]{ 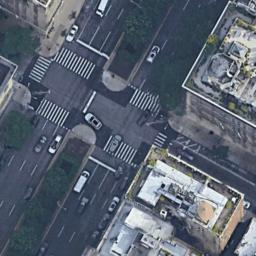}}
		\end{minipage}\vspace{-3mm}\hspace{-1mm}
        \end{subfigure}
        
  \centering
	\begin{subfigure}[t]{0.32\linewidth}
		\begin{minipage}[t]{\linewidth}
			\centering
			\raisebox{-0.15cm}{\includegraphics[width=1\linewidth]{ 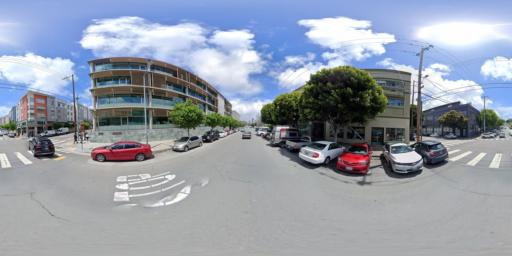}}
		\end{minipage}\hspace{-1mm}
	\end{subfigure}
        \begin{subfigure}[t]{0.16\linewidth}
		\begin{minipage}[t]{\linewidth}
			\centering
			\raisebox{-0.15cm}{\includegraphics[width=1\linewidth]{ 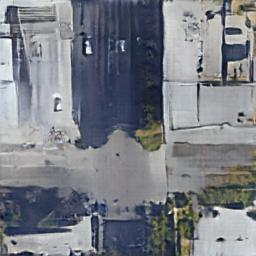}}
		\end{minipage}\hspace{-1mm}
	\end{subfigure}
        \begin{subfigure}[t]{0.16\linewidth}
		\begin{minipage}[t]{\linewidth}
			\centering
			\raisebox{-0.15cm}{\includegraphics[width=1\linewidth]{ 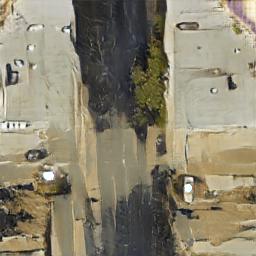}}
		\end{minipage}\hspace{-1mm}
	\end{subfigure}
        \begin{subfigure}[t]{0.16\linewidth}
		\begin{minipage}[t]{\linewidth}
			\centering
			\raisebox{-0.15cm}{\includegraphics[width=1\linewidth]{ 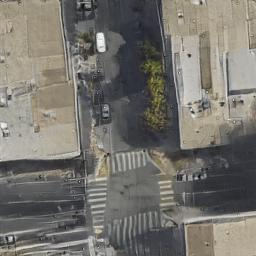}}
		\end{minipage}\hspace{-1mm}
        \end{subfigure}
        \begin{subfigure}[t]{0.16\linewidth}
        \begin{minipage}[t]{\linewidth}
			\centering
			\raisebox{-0.15cm}{\includegraphics[width=1\linewidth]{ 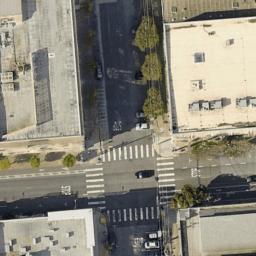}}
		\end{minipage}\vspace{-3mm}\hspace{-1mm}
        \end{subfigure}

    \centering
	\begin{subfigure}[t]{0.32\linewidth}
		\begin{minipage}[t]{\linewidth}
			\centering
			\raisebox{-0.15cm}{\includegraphics[width=1\linewidth]{ 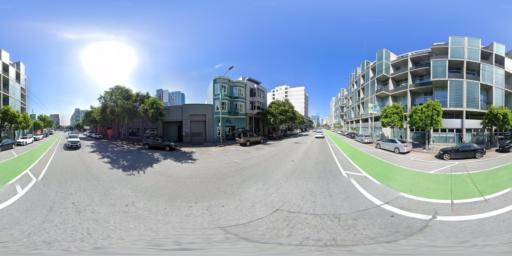}}
	\caption{Input}	
        \end{minipage}\hspace{-1mm}
        \end{subfigure}
	\begin{subfigure}[t]{0.16\linewidth}
		\begin{minipage}[t]{\linewidth}
			\centering
			\raisebox{-0.15cm}{\includegraphics[width=1\linewidth]{ 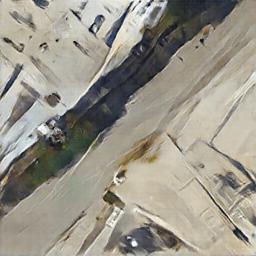}}
            \caption{Pix2Pix}
		\end{minipage}\hspace{-1mm}
	\end{subfigure}
	\begin{subfigure}[t]{0.16\linewidth}
		\begin{minipage}[t]{\linewidth}
			\centering
			\raisebox{-0.15cm}{\includegraphics[width=1\linewidth]{ 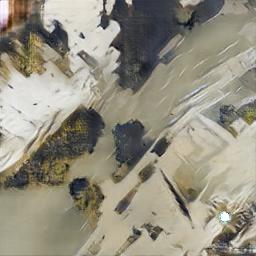}}
	\caption{CDE}	
        \end{minipage}\hspace{-1mm}
	\end{subfigure}
	\begin{subfigure}[t]{0.16\linewidth}
		\begin{minipage}[t]{\linewidth}
			\centering
			\raisebox{-0.15cm}{\includegraphics[width=1\linewidth]{ 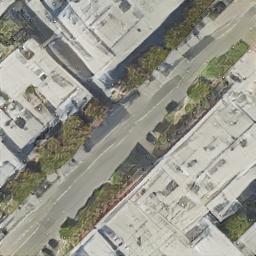}}
    \caption{Ours}		
        \end{minipage}\hspace{-1mm}
        \end{subfigure}
	\begin{subfigure}[t]{0.16\linewidth}
        \begin{minipage}[t]{\linewidth}
			\centering
			\raisebox{-0.15cm}{\includegraphics[width=1\linewidth]{ 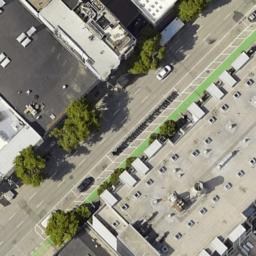}}
	\caption{GT}
        \end{minipage}\vspace{-3mm}\hspace{-1mm}
       \end{subfigure}
     
	\caption{Comparison of g2a (ground-to-aerial) synthesis on the VIGOR-GEN dataset.} 
%\vspace{-5mm}
 \label{VIGOR-GEN-g2a}
\end{figure} 

Further evidence supporting our idea is that when generating aerial-view images, other methods only produce blurred border regions. As demonstrated in Figure \ref{VIGOR-GEN-g2a}, Pix2Pix \cite{Pix2Pix} and CDE \cite{coming2021} generate central areas that are barely clear while introducing artifacts and blurs in the roof or non-central regions, where view-specific semantics aerial image information resides. 

\subsection{Ablation Study}

In this section, we perform ablation studies to validate the effectiveness of each component in our method.
We report variant models at the g2a direction on CVUSA.
As the key design of our method, we first replace the retrieval embedder with a trainable pix2pix encoder (\textbf{\romannumeral1}). 
% This means that the source images are encoded without identifying the shared information and closer in embedding space across two views.
% This means when encoding the source images, the shared information across two views is not identified and does not get closer to the embedding space.
% The embedding space is non-smooth 
In this way, it is difficult for the model to transform the information from the source view to the target view, as there still exists a large domain gap.

The second experiment removes the attn-AdaIN in our model (\textbf{\romannumeral2}). 
Without this component, the model loses the capability to fuse the retrieval embedding into the corresponding semantic regions effectively, which leads to a decrease in similarity.
% Such a modification loses the advantage of fusing retrieval embeddings and style information in deep layers, resulting in degradation in both realism and similarity.
% Such a modification loses the advantage of fusing retrieval embeddings and style information in deep layers, resulting in degradation in both realism and similarity.

\begin{table}
%\vspace{-8mm}
\caption{Ablation studies of our network on the CVUSA dataset.}
%\vspace{-2mm}
% \resizebox{\linewidth }{14mm}{
% \resizebox{\linewidth }{20mm}{
\centering
{\renewcommand\arraystretch{1}
 			\setlength{\tabcolsep}{1.2mm}{
    \begin{tabular}{l|ccccc}
    % {c|c|ccccc|ccccc}
        \bottomrule
         \multirow{2}{*}{\bf Method} & \multicolumn{5}{c}{\bf CVUSA}  \\ \cline{2-6} 
         ~&\makebox[0.04\textwidth][c]{\bf SSIM$\uparrow$}&\makebox[0.04\textwidth][c]{\bf PSNR$\uparrow$}&\makebox[0.04\textwidth][c]{\bf LPIPS$\downarrow$}&\makebox[0.04\textwidth][c]{\bf FID$\downarrow$}&\makebox[0.04\textwidth][c]{\bf R@1$\uparrow$} \\ \hline
         Ours & 0.3702  & 14.33 & \textbf{0.4302} & \textbf{13.57} & \textbf{96.25} \\
         (\textbf{\romannumeral1})w/o Embedder & 0.3312  & 13.66 & 0.4656 & 38.81 & 12.67 \\
         (\textbf{\romannumeral2})w/o Attn-AdaIN & 0.3629 & 14.01 & 0.4461 & 16.51 & 89.42 \\
         (\textbf{\romannumeral3})w/o Style & \textbf{0.3720} & \textbf{14.28} & 0.4412 & 17.88 & 94.23 \\
         (\textbf{\romannumeral4})w/o Ret. & 0.3571 & 13.75 & 0.4377 & 15.74 & 87.67 \\
         (\textbf{\romannumeral5})Same Structure & 0.3490& 14.06 & 0.4332 & 14.29 & 96.12 \\
         (\textbf{\romannumeral6})w/o coarse D & 0.3454 & 14.11 & 0.4308 & 13.67 & 95.61 \\  \toprule
        
    \end{tabular}}}
    %\vspace{-4mm}
    \label{tab:ablation}
\end{table}

Next, we also analyze the roles of the style (\textbf{\romannumeral3}) and retrieval embedding (\textbf{\romannumeral4}) in our generator. 
The fusion of retrieved information and style improves the network from two perspectives: correspondence and diversity. By fusing the retrieved information and style, we improve the model from two key retrieval-related perspectives: correspondence and diversity.
First, by fusing embeddings in deep layers, the model ensures the generation of semantically consistent representations in the target view against the visual difference. This improves cross-view retrieval performance by aligning the target image more closely with the source view. We observe a degradation of the performance in various metrics from Table \ref{tab:ablation}, especially in R@1 (with ~9\% drop).
Second, the additional style information promotes the diversity of visual features and enriches the visual representations, which facilitates the generation of view-specific semantics information in the target view. 
% The model without style information tends to generate images that are closer to the target view in pixel-wise while ignoring the quality.
This is crucial for generating distinct images in retrieval tasks, where diversity in output is essential for improving retrieval accuracy. However, while this integration results in a slight increase in SSIM, it leads to significant declines in LPIPS and FID scores.
% The retrieval embedding is also matter for generating the corresponding feature in the target view. 
% With fusing embeddings in deep layers, the model ensures the generation of semantically consistent representations in the target view to against the visual discrepancy.
% We observe a degradation of the performance in various metrics from Table \ref{tab:ablation}, especially in R@1 (with 5.58\% drop).
We further investigate the impact of different structural designs. If the model uses the same structure (i.e., ResBlock-S) to generate structural and facade information, metrics such as FID and LPIPS rise.
Besides, the performance of the model degrades if the discriminator for coarse images is disabled. 
Consequently, facade generation modules reinforce the performance of the network in cross-view synthesis.

\begin{figure}
% \vspace{-3mm}
\scalebox{.5}{
  \centering
		\begin{minipage}[t]{0.32\linewidth}
			\centering
			\raisebox{-0.15cm}{\includegraphics[width=1\linewidth]{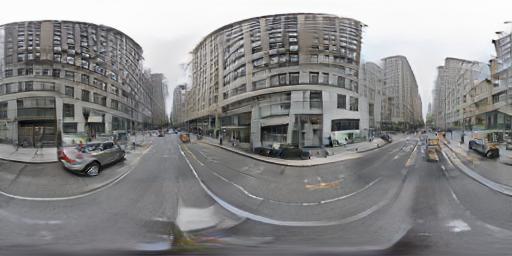}}
		\end{minipage}
		\begin{minipage}[t]{0.32\linewidth}
			\centering
			\raisebox{-0.15cm}{\includegraphics[width=1\linewidth]{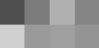}}
		\end{minipage}
		\begin{minipage}[t]{0.32\linewidth}
			\centering
			\raisebox{-0.15cm}{\includegraphics[width=1\linewidth]{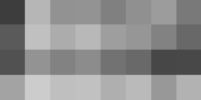}}
		\end{minipage}
		\begin{minipage}[t]{0.32\linewidth}
			\centering
			\raisebox{-0.15cm}{\includegraphics[width=1\linewidth]{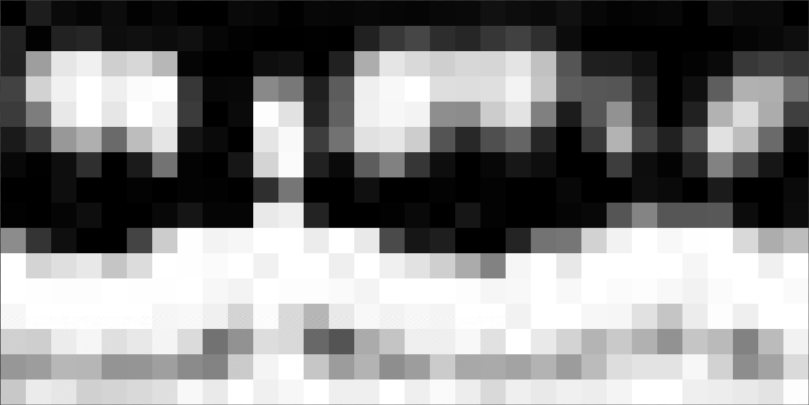}}
		\end{minipage}
        \begin{minipage}[t]{0.32\linewidth}
			\centering
			\raisebox{-0.15cm}{\includegraphics[width=1\linewidth]{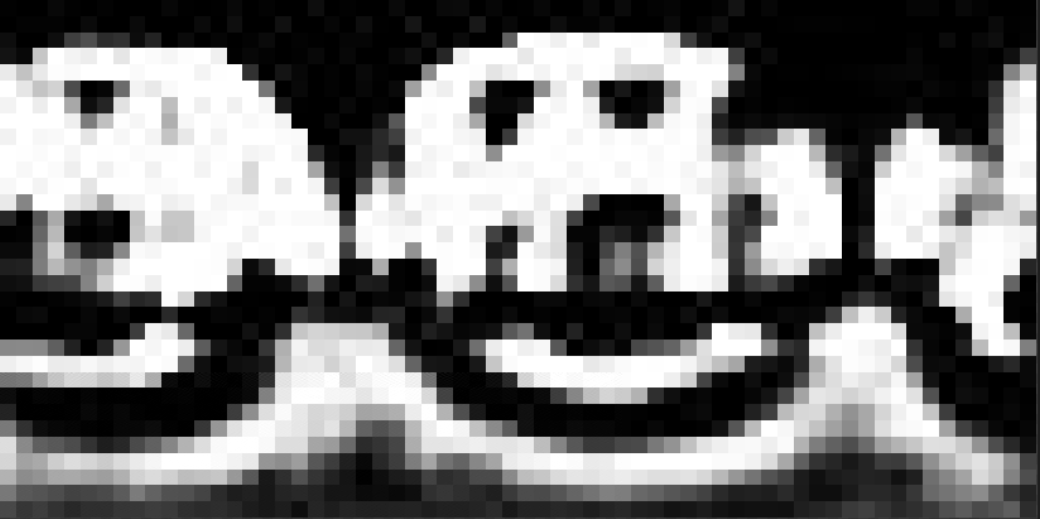}}
		\end{minipage}
        \begin{minipage}[t]{0.32\linewidth}
			\centering
			\raisebox{-0.15cm}{\includegraphics[width=1\linewidth]{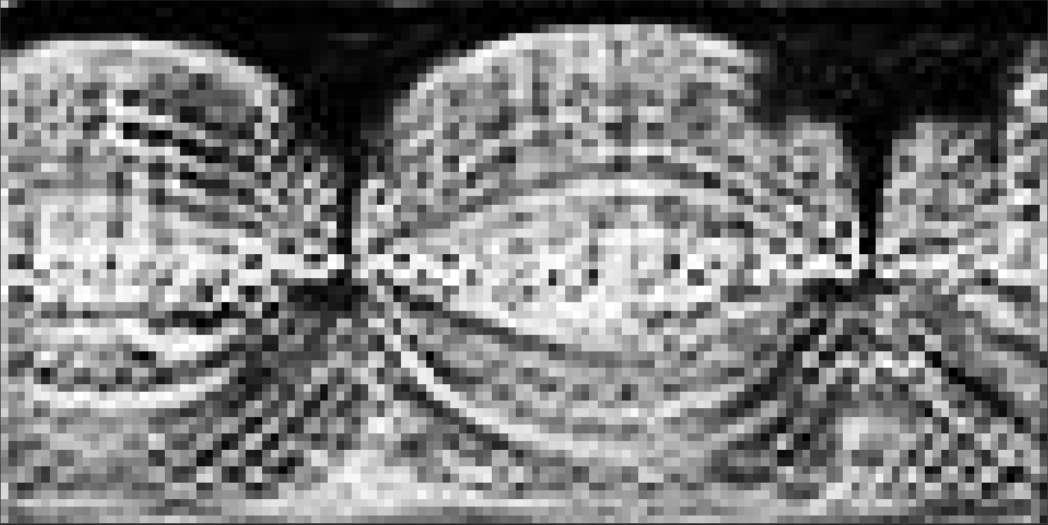}}
		\end{minipage}
  }
	\caption{Visualization of the weight map $M$ on VIGOR-GEN.} 
% \vspace{-4mm}
	\label{VIGOR-heatmap}
\end{figure}

Finally, to better understand the contribution of attn-AdaIN, we visualize the mask $M$ learned on different feature levels in Figure \ref{VIGOR-heatmap}, where the brighter pixel indicates the higher weight for retrieval embedding. This highlights the role of the retrieval embedding in guiding the generation process by focusing on relevant features, which is crucial for successful information retrieval.

% It suggests that on the low-level features, retrieval embedding dominates while style information merely effects on some irrelevant regions such as the inner side of the road. 
% Nevertheless, in the high-level features, the style begins to show more interest in the building, while retrieval embedding turns its effective region into the contours of the target image. 

% \paragraph{The effect of the CAN.} 

% With fusing embeddings in deep layers, the model ensures that shared information is smoothly transformed from source view to target view, which facilities the generation of the semantically consistent representations in the target view to against the visual discrepancy
% With fusing embeddings in deep layers, the model ensures the semantically consistent representations in the target view to against the visual discrepancy.
% which is beneficial to generate semantic consistent representation in the target view to against the visual discrepancy. 

\subsection{Further Discussion}

\begin{table}
    %\vspace{-4mm}
\caption{Comparison of different Embedders in our generator on CVUSA at a2g. }
 \vspace{-3mm}
    \centering
    {\renewcommand\arraystretch{1.0}
 			\setlength{\tabcolsep}{2.0mm}{
    \begin{tabular}{c|ccccc}
     \bottomrule
         \multirow{2}{*}{\bf Embedder} & \multicolumn{5}{c}{\bf CVUSA} \\ \cline{2-6}
         ~&{\bf SSIM$\uparrow$}&{\bf PSNR$\uparrow$}&{\bf LPIPS$\downarrow$}&{\bf FID$\downarrow$}&{\bf R@1$\uparrow$} \\ \hline
        SAIG & \textbf{0.3706} & \textbf{14.32} & \textbf{0.4302} & \textbf{13.57} & \textbf{96.25} \\
        LPN & 0.3559 & 13.89 & 0.4544 & 25.29 & 30.45 \\
         \toprule
    \end{tabular}}}
    \label{tab:EmbedderCVUSA}
    %\vspace{-3mm}
\end{table}

The retrieval embedder bridges the domain gap and provides a stable direction of gradient descent. By using retrieval loss, the embedding space remains smooth, aiding in better cross-view retrieval. Specifically, when the model generates an incorrect target image identity, the retrieval loss-guided embedding offers an optimal gradient direction for the generator to correct the identity. In contrast, embedders trained for discriminative tasks may produce non-smooth embedding spaces. For example, we compare the use of LPN \cite{wang2021LPN} in the generator, which treats cross-view image retrieval as a classification task and applies instance loss \cite{zheng2020university}. As shown in Table \ref{tab:EmbedderCVUSA}, the generator using LPN \cite{wang2021LPN} performs significantly worse than the generator using SAIG, highlighting the superiority of retrieval loss-based embedding.
% The embedder trained using retrieval loss is smooth in the embedding space. Once the model generates an incorrect identity of the target image, the embedding using retrieval loss can provide a good gradient direction for the
% generator to change the identity correctly.
% In another type of emb\ref{tab:EmbedderCVUSA}edder, which is trained on a discriminative task, the space can become non-smooth.
% Therefore, we compare the use of LPN \cite{wang2021LPN} in the generator, which regards cross\cite{wang2021LPN}-view image retrieval as a classification and thus applies instance loss \cite{zheng2020university}. 
% As shown in Table , the performance of the generator using LPN \cite{wang2021LPN} is significantly worse than the generator using SAIG.

\paragraph{Model Size}
%\subsection{Model Size}
% To better illustrate the overhead of our model, we show the comparison with different models in terms of model size and speed. 
% As illustrated in Table \ref{tab:modelsize}, our model has fewer parameters and faster inference compared with other methods.
Table \ref{tab:modelsize} compares our model with state-of-the-art methods in terms of model size (Params), inference speed (FPS), and generation quality (FID). Our model achieves the smallest size (25.9M), significantly smaller than others like SelectionGAN (58.3M) and PanoGAN (88.0M), making it lightweight and easier to deploy. Additionally, it also achieves the highest inference speed (39.2 FPS), far surpassing methods like SelectionGAN (18.0 FPS) and PanoGAN (19.1 FPS), demonstrating its suitability for real-time applications. Most importantly, our model achieves the best generation quality with an FID score of 13.57, significantly outperforming competitors like CDE (20.63) and S2SP (44.15). These results highlight the effectiveness of our lightweight design and optimized architecture, balancing efficiency and performance for practical applications.

\begin{figure}

    \includegraphics[scale=0.35]{ 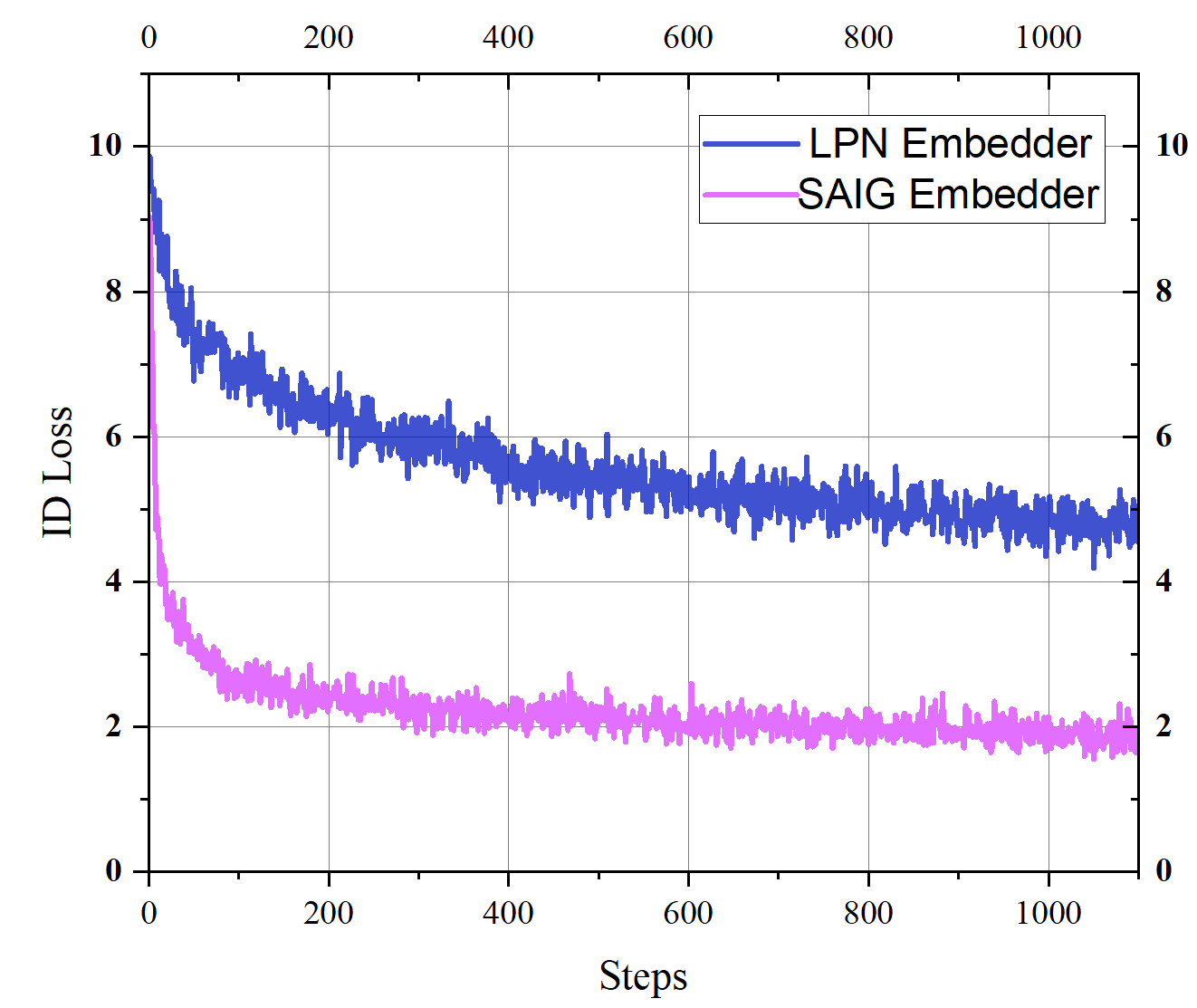}
    \caption{The curve of ID loss in generator using different embedders.}
    \label{fig:loss}
    \vspace{-4mm}
\end{figure}

\paragraph{Loss}
% Moreover, we compare the use of LPN \cite{wang2021LPN} in the generator, which regards a cross-view image retrieval as a classification and thus applies the instance loss \cite{zheng2020university}.
% This type of embedder, which is trained on a discriminative task, makes the space non-smooth.
% Although the classification embedder can also bridge the domain gap, its generation performance and convergence speed are significantly lower than that of the generator using the retrieval embedder, as shown in Table \ref{tab:EmbedderCVUSA} and Figure \ref{fig:loss}.
% As can be seen from Table \ref{tab:EmbedderCVUSA}, the performance of the generator using LPN as the embedder is poorer than that of the generator using SAIG, especially for R@1 (about \~66\% degradation).
The experiment investigates the impact of different embedders on the generator's performance and convergence, particularly comparing the SAIG embedder with the LPN embedder. The LPN embedder, as introduced in \cite{wang2021LPN}, treats cross-view image retrieval as a classification problem and employs instance loss \cite{zheng2020university} for training. However, this approach leads to a non-smooth embedding space, which negatively affects the generator's performance and convergence speed. As shown in Table \ref{tab:EmbedderCVUSA}, the generator with the SAIG embedder outperforms the LPN embedder across all metrics, including SSIM, PSNR, LPIPS, FID, and R@1, indicating superior generation quality and domain adaptation. Additionally, figure \ref{fig:loss} illustrates that the ID loss of the generator converges more rapidly and stably with the SAIG embedder compared to the LPN embedder, further highlighting the advantages of the retrieval-based embedding strategy in achieving smoother and more effective optimization.

\begin{figure*}

        \centering
		\begin{minipage}[t]{0.3\linewidth}
			\centering
			\raisebox{-0.15cm}{\includegraphics[width=1\columnwidth]{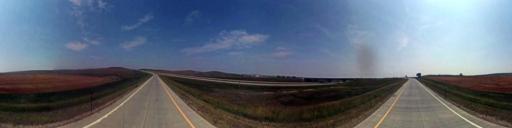}}
		\end{minipage}
		\begin{minipage}[t]{0.3\linewidth}
			\centering
			\raisebox{-0.15cm}{\includegraphics[width=1\columnwidth]{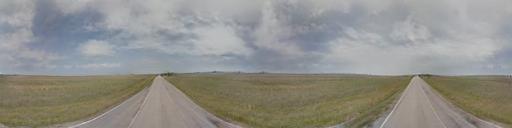}}
		\end{minipage}
		\begin{minipage}[t]{0.3\linewidth}
			\centering
			\raisebox{-0.15cm}{\includegraphics[width=1\columnwidth]{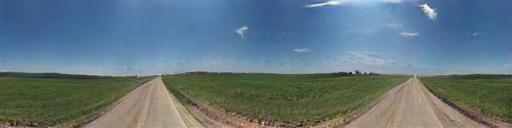}}
		\end{minipage}
`       
        \centering
		\begin{minipage}[t]{0.3\linewidth}
			\centering
			\raisebox{-0.15cm}{\includegraphics[width=1\columnwidth]{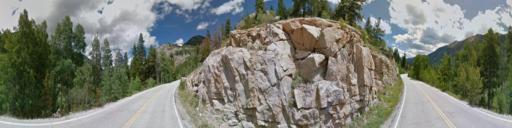}}
		\end{minipage}
		\begin{minipage}[t]{0.3\linewidth}
			\centering
			\raisebox{-0.15cm}{\includegraphics[width=1\columnwidth]{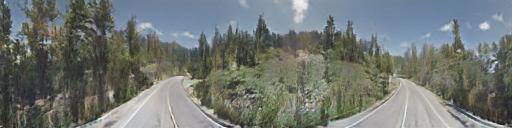}}
		\end{minipage}
		\begin{minipage}[t]{0.3\linewidth}
			\centering
			\raisebox{-0.15cm}{\includegraphics[width=1\columnwidth]{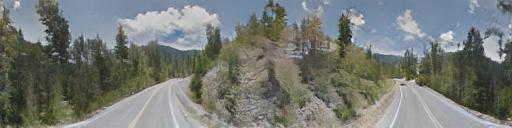}}
		\end{minipage}

        \centering
		\begin{minipage}[t]{0.3\linewidth}
			\centering
			\raisebox{-0.15cm}{\includegraphics[width=1\columnwidth]{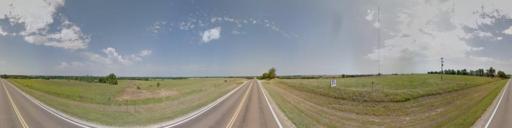}}
		\end{minipage}
		\begin{minipage}[t]{0.3\linewidth}
			\centering
			\raisebox{-0.15cm}{\includegraphics[width=1\columnwidth]{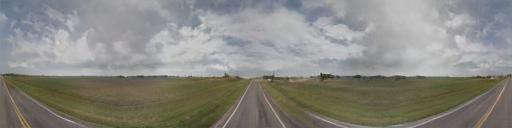}}
		\end{minipage}
		\begin{minipage}[t]{0.3\linewidth}
			\centering
			\raisebox{-0.15cm}{\includegraphics[width=1\columnwidth]{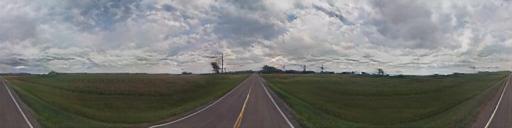}}
		\end{minipage}

        \centering
		\begin{minipage}[t]{0.3\linewidth}
			\centering
			\raisebox{-0.15cm}{\includegraphics[width=1\columnwidth]{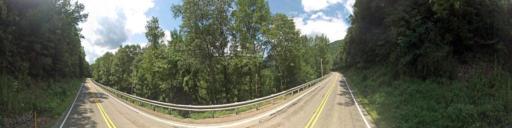}}
		\end{minipage}
		\begin{minipage}[t]{0.3\linewidth}
			\centering
			\raisebox{-0.15cm}{\includegraphics[width=1\columnwidth]{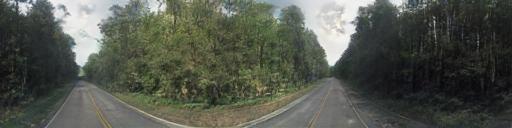}}
		\end{minipage}
		\begin{minipage}[t]{0.3\linewidth}
			\centering
			\raisebox{-0.15cm}{\includegraphics[width=1\columnwidth]{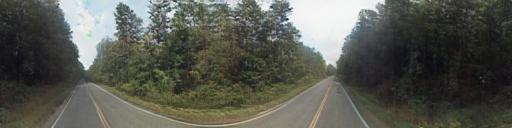}}
		\end{minipage}

	\centering
		\begin{minipage}[t]{0.3\linewidth}
			\centering
			\raisebox{-0.15cm}{\includegraphics[width=1\columnwidth]{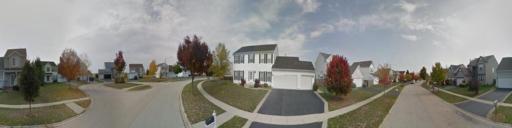}}
		\end{minipage} 
        \begin{minipage}[t]{0.3\linewidth}
        \centering
        \raisebox{-0.15cm}{\includegraphics[width=1\columnwidth]{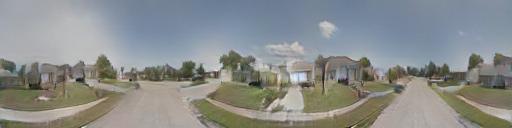}}
      \end{minipage}
      \begin{minipage}[t]{0.3\linewidth}
        \centering
        \raisebox{-0.15cm}{\includegraphics[width=1\columnwidth]{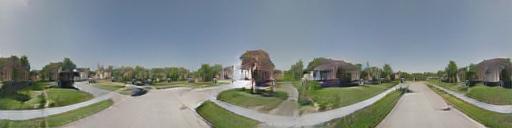}}
      \end{minipage}
	
	\centering
		\begin{minipage}[t]{0.3\linewidth}
			\centering
			\raisebox{-0.15cm}{\includegraphics[width=1\columnwidth]{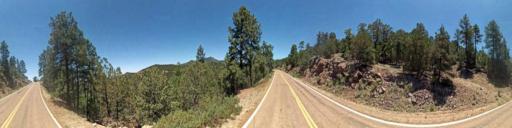}}
		\end{minipage}
    \begin{minipage}[t]{0.3\linewidth}
    \centering
    \raisebox{-0.15cm}{\includegraphics[width=1\columnwidth]{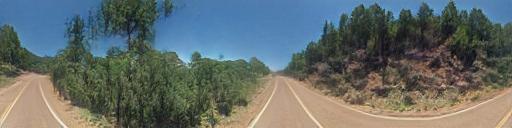}}
  \end{minipage}
  \begin{minipage}[t]{0.3\linewidth}
    \centering
    \raisebox{-0.15cm}{\includegraphics[width=1\columnwidth]{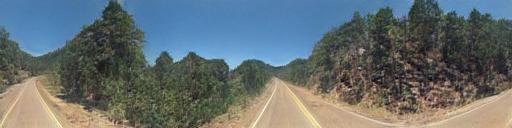}}
  \end{minipage}

  \centering
		\begin{minipage}[t]{0.3\linewidth}
			\centering
			\raisebox{-0.15cm}{\includegraphics[width=1\columnwidth]{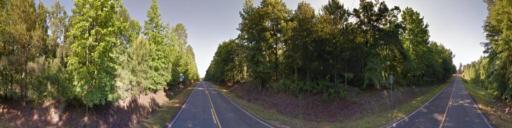}}
		\end{minipage}
    \begin{minipage}[t]{0.3\linewidth}
    \centering
    \raisebox{-0.15cm}{\includegraphics[width=1\columnwidth]{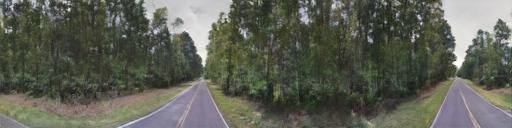}}
  \end{minipage}
  \begin{minipage}[t]{0.3\linewidth}
    \centering
    \raisebox{-0.15cm}{\includegraphics[width=1\columnwidth]{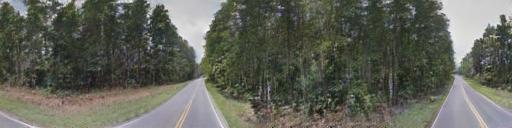}}
  \end{minipage}
  
  \centering
		\begin{minipage}[t]{0.3\linewidth}
			\centering
			\raisebox{-0.15cm}{\includegraphics[width=1\columnwidth]{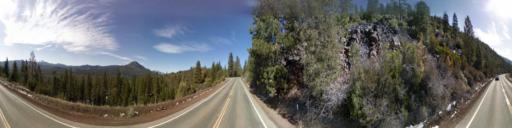}}
		\end{minipage}
    \begin{minipage}[t]{0.3\linewidth}
    \centering
    \raisebox{-0.15cm}{\includegraphics[width=1\columnwidth]{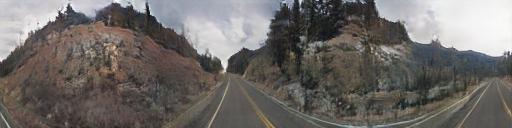}}
  \end{minipage}
  \begin{minipage}[t]{0.3\linewidth}
    \centering
    \raisebox{-0.15cm}{\includegraphics[width=1\columnwidth]{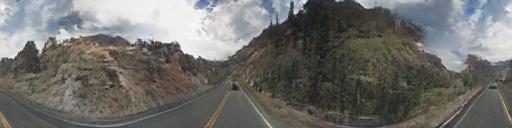}}
  \end{minipage}
    \centering
		\begin{minipage}[t]{0.3\linewidth}
			\centering
			\raisebox{-0.15cm}{\small{GT}}
		\end{minipage}
    \begin{minipage}[t]{0.3\linewidth}
    \centering
    \raisebox{-0.15cm}{\small{Structure-S}}
    \end{minipage}
  \begin{minipage}[t]{0.3\linewidth}
    \centering
    \raisebox{-0.15cm}{\small{Structure-S + Structure-T}}
  \end{minipage}
	\caption{Comparison of images generated by models using different Residual Blocks.} 
	\label{cvusa_same_struct}
\end{figure*}

\paragraph{Different Residual Block}
% In Figure \ref{cvusa_same_struct}, we show the effect of using different residual blocks in the experiments. It can be observed that the images generated using only Structure-S have more artifacts.
% This can be further illustrated by the ablation study in the main paper.
Based on the experimental results shown in Figure \ref{cvusa_same_struct}, it can be observed that models using only Structure-S generate images with visible artifacts, such as texture inconsistencies and less realistic details, particularly in complex scenarios like dense vegetation or architectural features. In contrast, the combination of Structure-S and Structure-T leads to substantial improvements, producing images that are smoother, more coherent, and closer to the ground truth. The Structure-S block captures global structural information effectively, while the addition of Structure-T enhances local detail modeling, balancing the global and local features. This combination results in sharper object contours, improved texture quality, and an overall more realistic appearance. These findings validate the benefits of combining both residual blocks and align with the ablation studies, demonstrating superior image quality and structural accuracy.

\begin{table}
 \caption{Comparison of model size for different model. }
    \centering
     {\renewcommand\arraystretch{0.8}
 			\setlength{\tabcolsep}{2mm}{
    \begin{tabular}{c|ccccc}
    \bottomrule
         \bf Model & \bf \#Params & \bf FPS & \bf FID \\ \cline{2-4}
        Pix2Pix & 41.8M & 34.1 &82.84\\ 
        XFork & 39.2M & 33.8 &79.75\\ 
        SelectionGAN & 58.3M & 18.0 & 90.72\\ 
        PanoGAN & 88.0M & 19.1 &75.24\\ 
        CDE & 37.3M & 35.9 & 20.63\\ 
        S2SP & 33.6M & 22.1 & 44.15\\
        Ours & \textbf{25.9M} & \textbf{39.2} & \textbf{13.57}\\  \toprule
    \end{tabular}}}
    \label{tab:modelsize}
\end{table}
%\vspace{-3mm}

\begin{table}
\caption{Comparison of our generator with different methods on CVUSA at $256\times1024$. }
    \centering
    {\renewcommand\arraystretch{1.0}
 			\setlength{\tabcolsep}{0.4mm}{
    \begin{tabular}{cl|ccccc}
    \bottomrule
         \bf Dataset&\bf Method&{\bf SSIM$\uparrow$}&{\bf PSNR$\uparrow$}&{\bf LPIPS$\downarrow$}&{\bf FID$\downarrow$}&{\bf R@1$\uparrow$} \\ \hline
        \multirow{3}{*}{CVUSA} & SelectionGAN & 0.4010 & 13.21 & 0.6169 & 103.27 & 3.81 \\
        ~&PanoGAN & 0.3575 & 13.47 & 0.5566 & 81.91 & 30.58 \\
        ~&Ours & \textbf{0.4232} & \textbf{14.11} & \textbf{0.4978} & \textbf{17.88} & \textbf{96.01} \\ \hline 
        \multirow{3}{*}{CVACT} & SelectionGAN & 0.4876 & 14.28 & 0.5232 & 97.63 & 5.76 \\
        ~&PanoGAN & 0.4915 & 14.31 & 0.4959 & 86.61 & 23.02 \\
        ~&Ours & \textbf{0.5513} & \textbf{14.48} & \textbf{0.4938} & \textbf{24.62} & \textbf{86.91 }\\ \hline 
        \multirow{3}{*}{VIGOR-GEN} & SelectionGAN & 0.4154 & 13.11 & 0.5225 & 106.24 & 7.80\\
        ~&PanoGAN & 0.4229 & 13.68 & 0.4933 & 79.72 &  8.26\\
        ~&Ours & \textbf{0.4771} & \textbf{14.01} & \textbf{0.4876} & \textbf{23.54} &  \textbf{36.18}\\  \toprule
    \end{tabular}}}
    \label{tab:hq}
\end{table}

\paragraph{Higher Quality}
% To better demonstrate the capability of our model, we perform higher resolution ($256\times1024 $) cross-view image synthesis. Compared to the existing methods (SelectionGAN \cite{tang2019multi} and PanoGAN \cite{wu2022cross}), our method performs better in all metrics, shown in Table \ref{tab:hq}.
% To showcase the superior capability of our model, we conduct experiments on high-resolution cross-view image synthesis at \(256 \times 1024\) resolution. As shown in Table \ref{tab:hq}, our method outperforms existing state-of-the-art approaches, including SelectionGAN \cite{tang2019multi} and PanoGAN \cite{wu2022cross}, across all evaluation metrics. Specifically, our model achieves significant improvements in SSIM, PSNR, LPIPS, FID, and retrieval accuracy (R@1), demonstrating its ability to generate higher-quality images with better structural consistency and perceptual fidelity. These results emphasize the effectiveness of our approach in bridging the domain gap and synthesizing realistic high-resolution images compared to previous methods.
To showcase the superior capability of our model, we conduct experiments on high-resolution cross-view image synthesis at \(256 \times 1024\)  resolution. As shown in Table \ref{tab:hq}, our method outperforms existing state-of-the-art approaches, including SelectionGAN \cite{tang2019multi} and PanoGAN \cite{wu2022cross}, across all evaluation metrics. Specifically, our model achieves significant improvements in SSIM, PSNR, LPIPS, FID, and R@1. The improved retrieval accuracy highlights our model’s ability to generate images that are more consistent and relevant in the context of cross-view image retrieval. The enhanced retrieval accuracy underscores our model's ability to generate images that are not only structurally consistent but also contextually relevant for cross-view image retrieval. These results demonstrate our approach excels in producing high-quality images with improved structural consistency and perceptual fidelity.

\section{Conclusion}

% In this work, we introduce a retrieval-guided solution for cross-view photo-realistic image synthesis. 
% % The synthesized images not only own content corresponding to the source view but also contain reasonably distinctive content of the target view.
% Specifically, we adopt a retrieval-guided framework that employs a retrieval network as the embedder and thus extracts information corresponding to the target view from the source images.
% % Such a framework is free of preprocessing and other additional information.
% Furthermore, we propose new generators to better generate structure and facade, which facilitates correspondence and the generation of view-specific semantics in the target view.
% In addition, we also build a large-scale, more practical, and challenging dataset (VIGOR-GEN) in the urban setting. 
% % Specifically, we offer the center-aligned image pairs for generation, and clean the unreasonable image pairs in the native dataset. 
% Through extensive experiments, it has been verified that our method outperforms other competitive methods.
%quantitative and qualitative

In this work, we present a retrieval-guided approach for cross-view  image synthesis, leveraging information retrieval techniques to bridge the domain gap in challenging cross-view tasks. By incorporating a retrieval network as the embedder, our framework effectively captures view-invariant and view-specific semantics, enabling accurate cross-view correspondence modeling without the need for auxiliary data. Additionally, we introduce novel generators that enhance the synthesis of both structure and facade, improving the generation of view-specific details in the target image. To facilitate model evaluation in real-world scenarios, we also introduce VIGOR-GEN, a large-scale, urban-focused dataset. Extensive experiments on CVUSA, CVACT, and VIGOR-GEN confirm the effectiveness of our method, achieving state-of-the-art performance across multiple evaluation metrics, including SSIM and FID. Our work successfully bridges the gap between information retrieval and image synthesis, offering valuable insights into how retrieval techniques can drive improvements in complex cross-domain generation tasks.

\bibliographystyle{ACM-Reference-Format}
\bibliography{sample}

\end{document}